\documentclass[journal]{IEEEtran}

\usepackage[colorlinks,linkcolor=red,anchorcolor=blue,citecolor=green]{hyperref}
\usepackage{graphicx}
\usepackage{dsfont}
\usepackage{stfloats}
\usepackage{subfigure}
\usepackage{amsmath}
\usepackage{cite}
\usepackage{booktabs}
\usepackage{algorithm}
\usepackage{algorithmic}
\usepackage{amssymb}

\begin{document}

\title{Unsupervised Deep Slow Feature Analysis for Change Detection in Multi-Temporal Remote Sensing Images}

\author{ Bo~Du,~\IEEEmembership{Senior~Member,~IEEE,}
	Lixiang~Ru,~\IEEEmembership{}
	Chen~Wu,~\IEEEmembership{Member,~IEEE,}
	and~Liangpei~Zhang,~\IEEEmembership{Fellow,~IEEE}
	\thanks{Manuscript submitted December 2, 2018, revised June 12, 2019. This work was supported in part by the National Natural Science Foundation of China under Grants 61601333, 61822113, and 41871243. \textit{Corresponding author: Chen Wu.}}
	\thanks{B. Du is with the School of Computer Science, and Collaborative Innovation Center of Geospatial Technology, Wuhan University, Wuhan, P.R. China (e-mail: gunspace@163.com). }
	\thanks{L. Ru is with the School of Computer Science, Wuhan University, Wuhan, P.R. China (e-mail:rulixiang@whu.edu.cn). }
	\thanks{C. Wu and L. Zhang are with the State Key Laboratory of Information Engineering in Surveying, Mapping and Remote Sensing, and School of Computer Science, Wuhan University, Wuhan, P.R. China (e-mail: chen.wu@whu.edu.cn, zlp62@whu.edu.cn).}
}

\markboth{IEEE Transactions on Geoscience and Remote Sensing}%
{Shell \MakeLowercase{\textit{et al.}}: Unsupervised deep slow feature analysis for change detection in multi-Temporal remote sensing images}

\maketitle

\begin{abstract}

	Change detection has been a hotspot in remote sensing technology for a long time. With the increasing availability of multi-temporal remote sensing images, numerous change detection algorithms have been proposed. Among these methods, image transformation methods with feature extraction and mapping could effectively highlight the changed information and thus has better change detection performance. However, changes of multi-temporal images are usually complex, existing methods are not effective enough. In recent years, deep network has shown its brilliant performance in many fields including feature extraction and projection. Therefore, in this paper, based on deep network and slow feature analysis (SFA) theory, we proposed a new change detection algorithm for multi-temporal remotes sensing images called Deep Slow Feature Analysis (DSFA). In DSFA model, two symmetric deep networks are utilized for projecting the input data of bi-temporal imagery. Then, the SFA module is deployed to suppress the unchanged components and highlight the changed components of the transformed features. The CVA pre-detection is employed to find unchanged pixels with high confidence as training samples. Finally, the change intensity is calculated with chi-square distance and the changes are determined by threshold algorithms. The experiments are performed on two real-world datasets {and a public hyperspectral dataset}. The visual comparison and quantitative evaluation have both shown that DSFA could outperform the other state-of-the-art algorithms, including other SFA-based {and deep learning methods.}

\end{abstract}


\IEEEpeerreviewmaketitle

\section{Introduction}

\IEEEPARstart{C}{hange} detection is defined as the process of identifying differences in the state of an object or phenomenon by observing it at different times \cite{singh1989review}. With the rapid development of remote sensing technology, more remote sensing images of the earth surface are now available \cite{du2013unsupervised,zhang2013tensor,du2012discriminative}. The multi-temporal remote sensing images covering the same area could help to detect land-cover and land-use changes, so that change detection could be better applied to diverse real-world applications, such as deforestation monitoring, damage assessment, vegetation phenology variation study, and disaster monitoring \cite{xian2009updating,xian2010updating,coppin1996digital,kennedy2009remote,wulder2008cross,luo2018urban}.

\par Generally, change detection algorithms could be divided into the following categories: 1) Image algebra methods mainly include image difference, image ratio, image regression, and change vector analysis \cite{ridd1998comparison,bruzzone2000automatic}. These methods directly calculate the difference between multi-temporal remote sensing images; 2) Image transformation algorithms extract the effective features of multi-temporal remote sensing images by transforming and combining their feature bands, and mainly include Principle Component Analysis (PCA) \cite{deng2008pca}, Multivariate Alteration Detection (MAD) \cite{nielsen1998multivariate,nielsen1997multivariate}, Gramm-Schmidt transformation (GS) \cite{collins1996assessment} and Independent Component Analysis \cite{marchesi2009ica}; 3) Classification methods mainly include post-classification and compound classification, which are both based on classification to obtain land-use categories \cite{bovolo2008novel,huang2010sampling,demir2012detection,ahlqvist2008extending}; 4) Other advanced methods contains the algorithms based on wavelet, Markov random field, and local gradual descent, etc. \cite{celik2011multitemporal,liu2008using,yetgin2012unsupervised,gueguen2011change}. Among all these kinds of change detection algorithms, image transformation methods have been widely studied and applied. The basic idea of image transformation is projecting the original multiband images into a new feature space to better separate changed and unchanged pixels. In this process, the most crucial work is to find an effective projecting algorithm to extract the determinative features.

\par Changed pixels in multi-temporal remote sensing images always have the feature differences with diverse change directions, while the features of unchanged pixels are supposed to be generally invariant \cite{singh1989review}. However, owing to the atmospheric conditions, illumination and sensor calibration and so on, those unchanged pixels always have slight differences \cite{coppin2004review,lu2004change}. Compared with changed pixels, changes of unchanged pixels usually have the consistent direction. By minimizing the feature variation of unchanged pixels, changed pixels could also be highlighted and separated. Inspired by this idea, slow feature analysis is proposed for detecting real changes and obtained satisfactory performance \cite{wu2014slow,wu2017post}.

\par SFA is a feature learning algorithm that extracts invariant and slowly varying features from input signals \cite{wiskott2002slow,wiskott2011slow}. And it has been successfully applied to solve diverse real-world problems, such as human action recognition, dynamic texture recognition and time series analysis, etc \cite{zhang2012slow,sun2014dl,franzius2008invariant,franzius2011invariant}. In change detection problems, changed and unchanged pixels correspond to quickly and slowly varying features in SFA, respectively. Based on this theory, \textit{Wu, et al.} \cite{wu2014slow} used SFA to suppress the spectral difference between slowly varying unchanged pixels, so that the changed pixels can be highlighted and well detected. By solving SFA problems, the proposed algorithms in \cite{wu2014slow} could get the projecting matrices to map original data, so that the unchanged components could be suppressed. All these algorithms have shown their good performance in some real-world remote sensing images. However, limited by the feature representative ability, linear SFA algorithms are sometimes not able to separate the changed and unchanged pixels. The potential solutions include projecting original feature into a higher-dimensional complex feature space to improve the model's complexity and feature representation ability.

\par {Actually, in \cite{wu2017kernel}, \textit{Wu, et al.} proposed a kernel version slow feature analysis (KSFA) for scene change detection. And the results have also shown that nonlinear extension of SFA is effective. However, in this method, KSFA is only designed for computing the change probabilities of bi-temporal scene level features. Some of its details are not suitable for pixel-wise change detection of multi-spectral imagery. Besides, KSFA is sensitive to the selection of kernel function. Different kernel function could lead to very different performances.}

\par Deep networks have been proved to have a powerful ability of representing non-linear functions, and thus can project original features into a more complex feature space \cite{bengio2007greedy,hinton2006reducing}. Due to the growing availability of both data and computing resources, deep neural networks have been resurging in these years. Numerous kinds of networks have been developed to complete different tasks, such as classification \cite{krizhevsky2012imagenet}, detection \cite{ren2015faster}, segmentation \cite{long2015fully}, and feature mapping\cite{hinton2006reducing}, etc. Besides, in recent years, deep networks have also been applied to learn non-linear transformations of highly correlated datasets, and performed well \cite{andrew2013deep}.

\par Therefore, inspired by the idea of utilizing deep network learning non-linear transformations, we propose a new algorithm called Deep Slow Feature Analysis (DSFA) in this paper. In DSFA, two deep networks are used to extract and represent the features of remote sensing images obtained at different times, respectively. The transformed features by deep networks are then taken as the inputs of SFA to obtain the projecting matrix. The projecting matrix could extract the most invariant component of multi-temporal remote sensing images, so the changed pixels could be accentuated. We formulate the loss function for DSFA model to make sure that the transformed features can represent the original data better. The intention of DSFA is to extract the invariant components of input features, which means that utilizing unchanged pixels as the inputs will help accelerating the training process and improving the final performance. However, in fact, labeled data are usually rare in remote sensing problems. Therefore, in DSFA, we use CVA to make a pre-detection and find unchanged pixel pairs as the inputs for training process. When the deep network is converged, the transformed features will be calculated by passing original features through trained networks. Then the difference of transformed features in SFA space is calculated. Finally, the change intensity map is calculated with chi-square distance, and the binary change map is obtained with threshold algorithms.
\par The rest of this paper is organized as follows. Section \ref{section2} introduces the SFA theory and the details of SFA in change detection. Section \ref{section3} presents the algorithm details of proposed DSFA. In Section \ref{section4}, we implement our proposed method and perform experiments on two real-world datasets and a public hyperspectral dataset. {In Section \ref{discussion}, some settings of our experiments are discussed.} And Section \ref{section5} draws the conclusion of this paper.

\section{Slow Featue Analysis}\label{section2}
\par In this section, we'll introduce the mathematical theory of SFA, and how SFA is extended to solve change detection problems. Mathematically, SFA is formulated as follows:

\par Given a multi-dimensional temporal signal $ s(t)=[s_1 (t),s_2 (t),\cdots,s_n (t)] $, where $n$ represents the dimension and $ t\in[t_0,t_1] $, the target of SFA is finding a set of transforming functions $[ g_1 (x),g_2 (x),\cdots ,g_M (x)]$ to generate the output signal $ z(t)=[g_1 (s),g_2 (s),\cdots,g_M (s)] $ and ensuring that transformed signal is time invariant as possible. Mathematically, the objective function of SFA is
\begin{equation}
	min_{g_j}: {\langle (\dot g_j(s))^2 \rangle_t, j\in[1,2,\cdots,M]},
\end{equation}
under the following constraints:
\begin{equation}
	\langle g_j (s)\rangle_t=0,
\end{equation}
\begin{equation}
	\langle g_j (s)^2 \rangle _t=1,
\end{equation}
\begin{equation}
	\forall i<j: \langle g_i (s) g_j (s)\rangle _t=0,
\end{equation}
where $\langle g_j (s)\rangle_t$ denotes the mean signal of $g_j (s)$ over time $t$ and $ \dot g_j (s)$ is the first-order derivate of $g_j (s)$. Therefore, the objective of SFA is minimizing the mean value of the first-order derivate of transformed signal. Among these constraints, Constraint (2) is to simplify the process of solving the optimization problem. Constraint (3) ensures that each output signal could contain certain information. And Constraint (4) is presented to eliminate the correlation between output signals and force each signal carries different type of information.
\par In the linear case, the transforming function could be expressed as a mapping matrix:
\begin{equation}
	g_j (s)=w_j^T s,
\end{equation}
where $w_j^T$ denotes the transposition of $w_j$. And the objective function and constraints could be reformulated as follows:
\begin{equation}
	\langle(w_j^T \dot s)^2 \rangle_t=w_j^T \langle \dot s \dot s ^T \rangle _t w_j =w_j^T A w_j,
\end{equation}
\begin{equation}
	\langle(w_j^T s)\rangle_t=0,
\end{equation}
\begin{equation}
	\langle(w_j^T s)(w_j^T s)\rangle_t=w_j^T \langle ss^T \rangle _t w_j=w_j^T Bw_j=1,
\end{equation}
\begin{equation}
	\langle(w_i^T s)(w_j^T s)\rangle_t=w_i^T \langle ss^T \rangle_t w_j=w_i^T Bw_j=0.
\end{equation}
\par In (6), $A=\langle \dot s \dot s ^T \rangle _t$ is the expectation of the covariance matrix of the first-order derivative of input signals. (7) represents Constraint (2), and it can be implemented by pre-processing the input data. (8) and (9) denote Constrain (3) and (4), respectively. And $B=\langle s s ^T \rangle _t$ is the expectation of covariance matrix of original input signals.

\par In SFA theory, (9) can be integrated to (6) as follows:
\begin{equation}
	\langle(w_j^T \dot s)^2 \rangle_t=w_j^T A w_j=\frac{w_j^T A w_j}{w_j^T B w_j}=\frac{\langle (w_j^T \dot s)^2 \rangle_t}{\langle(w_j^T s)(w_j^T s)\rangle_t}.
\end{equation}
\par And this optimization problem can be solved by the generalized eigenvalue problem:
\begin{equation}
	AW=BW\Lambda,
\end{equation}
where $W$ and $\Lambda$ is the generalized eigenvector matrix and a diagonal matrix of eigenvalues, respectively. According to (10) and (11), the most invariant
component of the output signal has the smallest eigenvalue.

\begin{figure*}[htbp]
	\centering
	\includegraphics[scale=1.05]{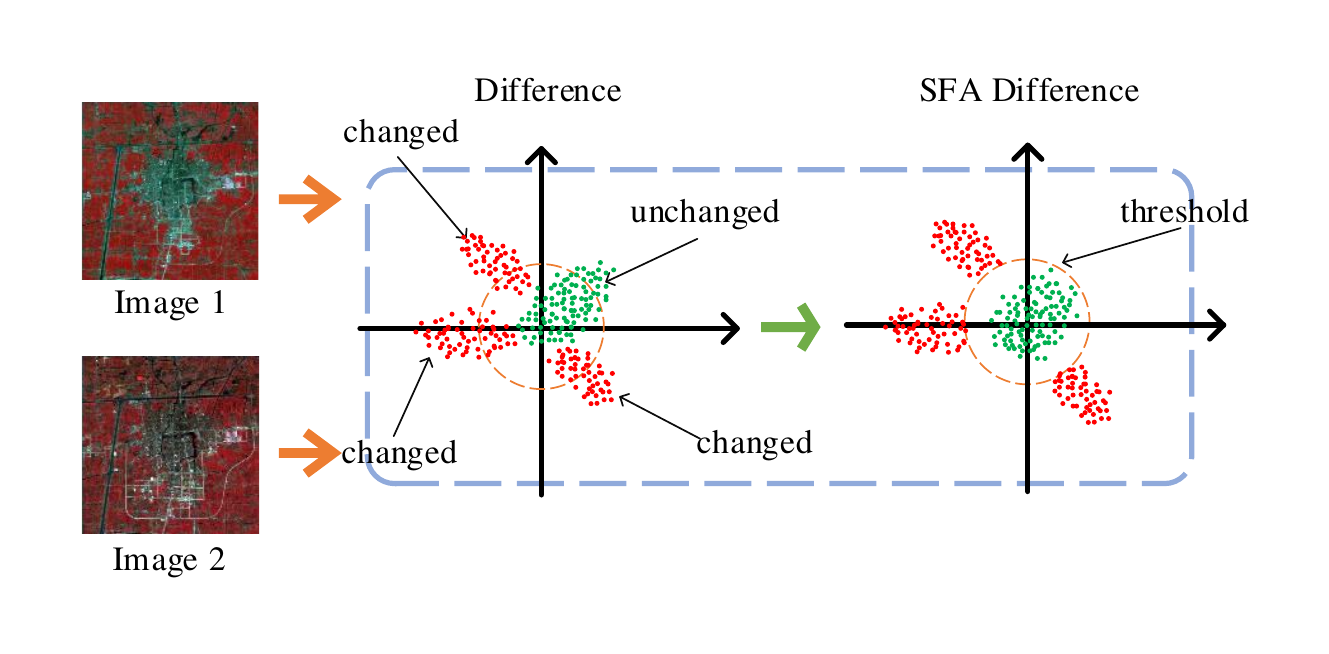}
	\caption{A schematic of SFA in change detection.}
	\label{sfa}
\end{figure*}

\par In pixel-based change detection problems, the input signals are raw pixels of remote sensing images, which are discrete. In consequence, SFA need to be reconstructed to cope with discrete cases. As shown in Figure~\ref{sfa}, the objective of SFA in change detection problems is suppressing unchanged pixels to highlight changed ones, so that they could be separated much easier. Mathematically, let $x_i,y_i\in \mathds{R}^m$ denote corresponding pixels in bi-temporal remote sensing images, where $m$ is the number of bands. After normalizing the input data, the objective of SFA is reformulated as
\begin{equation}
	min_{w_j}: \frac{1}{n} \sum_{i=1}^{n} (w^T_j x_i-w^T_j y_i )^2,
\end{equation}
where $n$ is the total number of pixels. And constraints are rewritten as
\begin{equation}
	\frac{1}{2n} [\sum_{i=1}^n w^T_j x_i+\sum_{i=1}^nw^T_j y_i]=0,
\end{equation}
\begin{equation}
	\frac{1}{2n} [\sum_{i=1}^n (w^T_j x_i)^2+\sum_{i=1}^n (w^T_j y_i )^2 ]=1,
\end{equation}
\begin{equation}
	\frac{1}{2n} [\sum_{i=1}^n (w^T_j x_i)(w^T_l x_i)+\sum_{i=1}^n (w^T_j y_i)(w^T_l y_i)]=0.
\end{equation}
\par In the generalized eigenvalue problem of SFA, $A$ and $B$ in $(11)$ are reformulated as follows:
\begin{equation}
	A=\frac{1}{n} \sum_{i=1}^n (x_i-y_i ) (x_i-y_i )^T,
\end{equation}
\begin{equation}
	B=\frac{1}{2n} [\sum_{i=1}^n x_i x_i^T +\sum_{i=1}^n y_i y_i^T ].
\end{equation}
\par When $A$ and $B$ are obtained, the eigenvector matrix $W$ will be solved. By normalizing $W$, the final mapping matrix is obtained.
\begin{equation}
	\hat w_j =\frac{w_j}{\sqrt{w^T_j Bw_j}}.
\end{equation}
\par Then the change detection result, the difference between transformed bi-temporal images, is calculated as $D_j=\hat w ^T x_j-\hat w^T y_j$.

\section{Methodology}\label{section3}
\par
\begin{figure*}[htbp]
	\centering
	\includegraphics[scale=1.15]{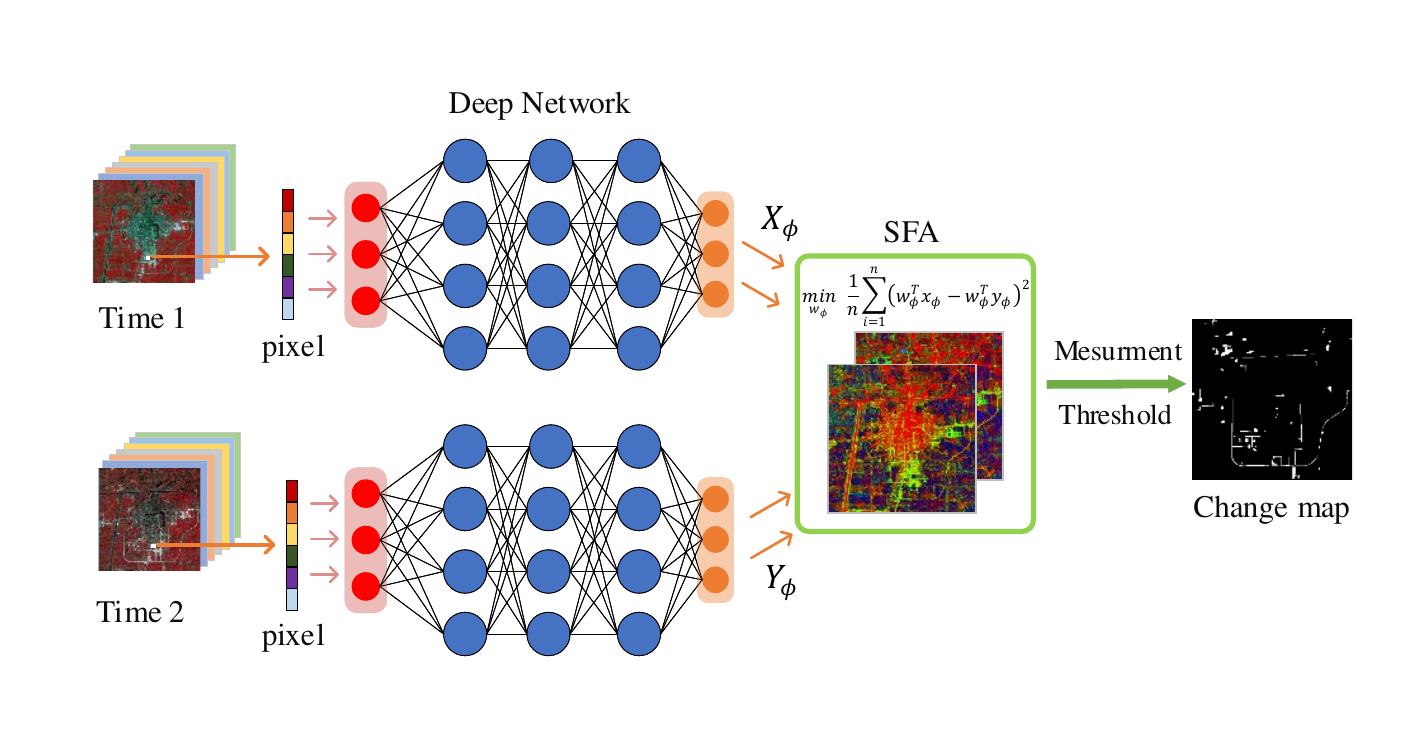}
	\caption{A schematic of DSFA, consisting two deep networks.}
	\label{dsfa}
\end{figure*}

 As mentioned above, those existing SFA-based change detection algorithms are all linear. In order to improve the representing ability of features and final change detection performance, in this section, we propose Deep Slow Feature Analysis (DSFA). The main structure of DSFA is shown in Figure~\ref{dsfa}.

\par As we can see in Figure~\ref{dsfa}, the input of DSFA is pairwise pixels of multi-temporal imagery. Then DSFA could be roughly divided to two parts: Deep Network module and SFA constraint. In the Deep Network module, two symmetric networks, whose layers are all Fully Connected Layer, are used to project original input data into a new complex high-dimensional feature space. In Figure~\ref{dsfa}, the red nodes denote the nodes of input layers, the blue nodes represent the nodes of hidden layers and the yellow nodes are used to represent output layers. Each hidden layer of the Deep Network module has the same number of nodes. After the original data is transformed, we use the SFA constraint to suppress the invariant components and highlight the changed components of transformed features. We formulate the loss function of DSFA so that the parameters of deep networks could be solved based on gradient-based optimization algorithms.

\subsection{Formulation}

\par Mathematically, DSFA is defined as follows: Assuming the original bi-temporal remote sensing images are $X, Y\in \mathds{R}^{m \times n}$, where $m$ and $n$ respectively denote the number of feature bands and pixels. For clarity, let $h_i$ denotes the number of nodes of the $i-th$ hidden layer of the networks, and $o$ is the number of nodes of the output layer. Given an instance $X$, the output of the first hidden layer could be formulated as
\begin{equation}
	f^1_1 (X)=s(w_1^1 X+b_1^1 ),
\end{equation}
where $w_1^1\in \mathds{R}^{h_1 \times m}$ and $b_1^1\in\mathds{R}^{h_1}$ denote the weight matrix and the bias vector, respectively. And $s(\cdot)$ represents the activation function. The output of the subsequent layers is calculated in the same way. For a network with $l$ hidden layers, the output of the last hidden layer is $f^1_l (X)=s(w_l^1 f^1_{l-1} (X)+b_l^1 $), where $w_1^l\in \mathds{R}^{h_l \times h_{l-1}}$ and $b_1^l\in \mathds{R}^{h_l}$. After that, $f^1_l (X)$ will be mapped by the output layer.

\par Finally, the final transformed feature of this network is
\begin{equation}
	X_\phi =f(\theta_1,X)=s(w_o^1 f^1_l (X)+b_o^1 ),
\end{equation}
where $w_o^1\in \mathds{R}^{o \times h_l}$ and $b_o^1\in \mathds{R}^o$ are the weight matrix and bias vector, respectively. And $\theta_1$ is the set of all the parameters in the network, including ${w_1^1,\cdots,w_l^1,w_o^1}$ and ${b_1^1,\cdots,b_l^1,b_o^1}$. And for another instance $Y$, $Y_\phi$ has a symmetric expression and meaning.
\begin{equation}
	Y_\phi=f(\theta_2,Y)=s(w_o^2 f^2_l (X)+b_o^2 ).
\end{equation}
\par When the original given data is mapped into a new high dimensional feature space by deep networks, let $\hat X_\phi= X_\phi - \frac{1}{n} \mathbf{1} X_\phi $ and $\hat Y_\phi= Y_\phi - \frac{1}{n} \mathbf{1} Y_\phi $ denote the centralized $X_\phi$ and $Y_\phi$, respectively, where $\mathbf{1}\in \mathds{R}^{o\times o}$ is a matrix whose elements are all 1. Then the covariance matrix of transformed data will be calculated.
\begin{equation}
	\Sigma_{XX}=\hat X_\phi \hat X_\phi ^T+r*I,
\end{equation}
\begin{equation}
	\Sigma_{YY}=\hat Y_\phi \hat Y_\phi ^T+r*I,
\end{equation}
\begin{equation}
	\Sigma_{XY}=(\hat X_\phi-\hat Y_\phi )(\hat X_\phi-\hat Y_\phi)^T.
\end{equation}
where $I$ denotes the identity matrix and {$r$ is a regularization constant. Assume that $r>0$, so that $\Sigma_{XX}$ and $\Sigma_{YY}$ are both positive definite and invertible. } Therefore, in DSFA problem, the generalized eigenvalue problem to be solved is formulated as:
\begin{equation}
	A_\phi W=B_\phi W\Lambda \Leftrightarrow B_\phi ^{-1} A_\phi W=W\Lambda,
\end{equation}
where $A_\phi=\Sigma_{XY}$ and $B_\phi=\frac{1}{2} (\Sigma_{XX}+\Sigma_{YY})$. According to $(22-24)$, the final form of this problem is
\begin{equation}
	[\frac{1}{2} (\Sigma_{XX}+\Sigma_{YY} )]^{-1} \Sigma_{XY} W=W\Lambda.
\end{equation}
\par Based on SFA theory, the most invariant component has the smallest eigenvalue. Thus, the objective of DSFA could be designed as minimizing the total square of all eigenvalues, so that the variance of unchanged pixels can be suppressed and changed pixels are much easier to be detected. The loss function of DSFA then could be formulated as follows:
\begin{equation}
	\mathcal{L}(\theta_1,\theta_2)=tr[(B_\phi^{-1} A_\phi )^2 ],
\end{equation}
where $tr(\cdot)$ denotes the trace of a matrix. Utilizing (27), the loss value of DSFA could be calculated and the parameters of networks $\theta_1$ and $\theta_2$ can be obtained with gradient-based optimization algorithm.

\subsection{Optimization}
\par To calculate the gradient of $\mathcal{L}(\theta_1,\theta_2)$ with respect to all the $w_l^v$ and $b_l^v$, we could use the back-propagation algorithm, which requires computing the gradient of $\mathcal{L}(\theta_1,\theta_2)$ with respect to $\hat X_\phi$ and $\hat Y_\phi$.

\par According to the reference \cite{IMM2012-03274}, and using the fact that $A_\phi$ and $B_\phi$ are both symmetric, we could then have:
\begin{equation}
	\nabla_A=\frac{\partial\mathcal{L}(\theta_1,\theta_2)}{\partial A_\phi}=2B^{-1}_\phi A_\phi B^{-1}_\phi ,
\end{equation}
\begin{equation}
	\nabla_B=\frac{\partial\mathcal{L}(\theta_1,\theta_2)}{\partial B_\phi}=-2B^{-1}_\phi A_\phi B^{-1}_\phi A_\phi B^{-1}_\phi.
\end{equation}
\par Utilizing the derivation in \cite{andrew2013deep}, we could have the gradient of $A_\phi$ with respect to each element of $\hat X_\phi$:
\begin{equation}
	\begin{split}
		\frac{\partial A_\phi^{ab}}{\partial \hat X_\phi^{ij}} &=\frac{1}{n}(\xi_{(a=i)}\hat X_\phi^{bj}+\xi_{(b=i)}\hat X_\phi^{aj})\\
		&-\frac{1}{n}(\xi_{(a=i)}\hat Y_\phi^{bj}+\xi_{(b=i)}\hat Y_\phi^{aj}),
	\end{split}
\end{equation}
where $\xi_{(e)}$ represents the indicator function. If $e$ is true, then $\xi_{(e)}=1$, otherwise $\xi_{(e)}=0$. Similarly, the gradient of $B_\phi$ with respect to each element of $\hat X_\phi$ is computed as follows:
\begin{eqnarray}
	\frac{\partial B_\phi^{ab}}{\partial \hat X_\phi^{ij}} =\frac{1}{2n}(\xi_{(a=i)}\hat X_\phi^{bj}+\xi_{(b=i)}\hat X_\phi^{aj}),
\end{eqnarray}
\par Integrating (28)-(31), the gradient of $\mathcal{L}(\theta_1,\theta_2)$ with respect to $\hat X_\phi^{ij}$ is:
\begin{equation}
	\begin{split}
		\frac{\partial\mathcal{L}(\theta_1,\theta_2)}{\partial\hat X_\phi^{ij}} &=\sum_{ab}\nabla_A^{ab}\frac{\partial A_\phi^{ab}}{\partial \hat X_\phi^{ij}}+\sum_{ab}\nabla_B^{ab}\frac{\partial B_\phi^{ab}}{\partial \hat X_\phi^{ij}} \\ &=\frac{2}{n}(\nabla_A\hat X_\phi-\nabla_A\hat Y_\phi)_{ij}+\frac{1}{n}(\nabla_B\hat X_\phi)_{ij}.
	\end{split}
\end{equation}
\par The derivation process isn't straight and its details are presented in Appendix \ref{appendix}. Finally, it's obvious that the gradient of $\mathcal{L}(\theta_1,\theta_2)$ with respect to $\hat X_\phi$ could be computed as:
\begin{equation}
	\frac{\partial\mathcal{L}(\theta_1,\theta_2)}{\partial\hat X_\phi}=\frac{2}{n}(\nabla_A\hat X_\phi-\nabla_A\hat Y_\phi)+\frac{1}{n}\nabla_B\hat X_\phi.
\end{equation}
\par And for another instance $Y_\phi$, the expression of $\mathcal{L}(\theta_1,\theta_2)/\partial Y_\phi$ is symmetric. We then could utilize Gradient Descent algorithms to minimize the loss to obtain the parameters of deep network module of DSFA.
\par According to loss function, the objective of DSFA is projecting the difference of pairwise pixels into an invariant difference feature space. Therefore, if we utilize unchanged pairwise pixels as training samples, the learned non-linear projection of deep network will have better performance in extracting the invariant components. However, in practice, priori labeled information in change detection is always hard to get. To select unchanged pairwise pixels for training process, in this paper, we use the CVA method to make a pre-detection. {In this process, CVA and Kmeans method are employed to obtain the difference map and the binary change map of input multi-temporal imagery, respectively. Training samples are then randomly selected from the detected unchanged areas}.
\par After obtained the training set and trained the network, the original data will be passed through the deep network to get the transformed features $X_\phi$ and $Y_\phi$. Then, the generalized eigenvalue problem will be solved to obtain the projecting matrix $w_\phi$ and the difference between mapped features is calculated as follows:
\begin{equation}
	D_\phi=w_\phi ^T X_\phi-w_\phi ^T Y_\phi.
\end{equation}
\par Then the change intensity of bi-temporal images could be calculated. In order to eliminate the differences in the scale of each feature bands, in this paper, we use chi-square distance to measure the intensity of changes, which is calculated as
\begin{equation}
	chi2=\sum_{i=1}^m \frac{(D_\phi^i )^2}{\sigma_i^2}.
	\label{chi2}
\end{equation}
\par In (35), $m$ is the number of feature bands, and $\sigma^2$ is variance of each bands {obtained by statistically analyzing}. Threshold algorithms, such as OSTU method and Kmeans method, are then employed to get the final binary change map. The whole detailed process of training and generating binary change map for DSFA is summarized in Algorithm \ref{alg:DSFA}.

\begin{algorithm}[htbp]
	\caption{Process of training and generating binary change map for DSFA.}
	\label{alg:DSFA}
	\begin{algorithmic}[1]
		\REQUIRE~~\\
		Multi-temporal input images $I^1$ and $I^2$;\\
		\ENSURE ~~\\
		The binary change map $D$;\\
		\STATE Standardize $I^1$ and $I^2$ using $z-score$ method;
		\STATE Employ CVA pre-detection to generate training samples $X$ and $Y$;
		\STATE Initialize the network's parameters \{$\theta_1, \theta_2$\};
		\label{code:fram:Initialize}
		\WHILE {$i < max\_epoches$}
		\STATE Calculate the projected features $X_\phi = f_1(X, \theta_1)$ and $ Y_\phi = f_2(Y, \theta_1);$
		\STATE Calculate the loss value:\\ $$\mathcal{L}(\theta_1, \theta_2)= tr[(B_\phi^{-1} A_\phi)^2];$$
		\STATE Calculate the gradient: $\partial \mathcal{L}(\theta_1, \theta_2)/\partial \theta_1$ and $ \partial \mathcal{L}(\theta_1, \theta_2)/\partial \theta_2;$
		\STATE Updating the parameters using Gradient Descent algorithm;
		\STATE $i$++;
		\ENDWHILE
		\label{code:fram:while}
		\STATE Calculate the mapped features $ I_\phi^1 $ and $ I_\phi^2 $ of $I^1$ and $ I^2 $;
		\STATE Solve SFA problem to obtain projecting matrix $w_\phi$;
		\label{code:fram:mapped}
		\STATE Calculate the difference map: \\ $$\Delta I =w_\phi^T I_\phi^1 - w_\phi^T I_\phi^2;$$
		\STATE Threshold to get the binary change map $D$;
		\label{code:fram:Threshold}
		\RETURN $D$;
	\end{algorithmic}
\end{algorithm}

\section{Experiment}\label{section4}
\par To evaluate the performance of DSFA, in this section, we implement DSFA on TensorFlow and perform experiments on three multi-temporal remote sensing image datasets. Datasets used in our experiment include two Enhanced Thematic Mapper (ETM) datasets {and a public hyperspectral change detection dataset}. The first one is Taizhou dataset, covering the city of Taizhou, China, acquired in 2000 and 2003. And the second is Nanjing dataset, which are respectively acquired in 2000 and 2002. Both datasets were obtained by the Landsat 7 Enhanced Thematic Mapper Plus (ETM+) sensor with a spatial resolution of 30 m. And 6 spectral bands (1-5 and 7) are selected for our experiments.{The band 6 has a spatial resolution of 60m, so it's dropped and not used in our experiments. The third dataset is River dataset \footnote{Avaliable: http://crabwq.github.io/}, and consists of two hyperspectral images with a size of $463\times241$, which are respectively obtained in May, 2013 and December, 2013, Jiangsu Province, China. Each image in this dataset contains 198 spectral bands after noisy bands removal.}

\subsection{Experiment settings}
\par In the DSFA model, the weight and bias matrices of each layer are initialized randomly, and need to be optimized. The other values, including the number of layers and nodes in each view and the DSFA regularization parameter in (22-24) are hyperparameters. {As for the DSFA regularization parameter, we tuned it over the range $[10^{-8},10^{-1}]$, and eventually selected $10^{-4}$ as the value for our proposed model. The influence of the regularization parameter $r$ is discussed in the Section \ref{discussion}.}
\par Some other conventional and SFA-based change detection algorithms are also implemented for comparison, including CVA, PCA \cite{deng2008pca}, MAD \cite{nielsen1998multivariate}, IRMAD \cite{nielsen2007regularized}, USFA \cite{wu2014slow}, ISFA \cite{wu2014slow}, PCANet \cite{Gao2016} and SDPCANet \cite{Li2019}. All of them are unsupervised algorithms. Before calculating the difference map, PCA uses Principal Component Analysis method to project original data into a new lower dimensional feature space. MAD is a change detection method based on the established theory Canonical Correlation Analysis (CCA), which is firstly proposed in \cite{hardoon2004canonical}. It utilizes CCA to maximize the correlation between the features of multi-temporal images. IRMAD is an iteratively weighted extension of MAD. It firstly calculates the original MAD variates. And in the following iterations, it applies different weights to each pixels or regions to emphasize the changed parts of images. USFA and ISFA are proposed in \cite{wu2014slow}. Based on the SFA theory, USFA computes a projecting matrix to suppress the unaltered components of input data to highlight changed components. And ISFA is an iteratively weighted extension of USFA, and has the same way to calculate weights as IRMAD. {PCANet method firstly takes gabor wavelets and fuzzy c-means as the pre-detection method to select the training samples. Then, a PCANet \cite{Chan2015} model is trained with the image patches centered at the interested pixels. Finally, the change map is obtained by classifying the remain patches with the trained model. SDPCANet developed PCANet by using a context-aware saliency detection method \cite{goferman2011context} to select more robust and confident training samples in the pre-detection process.}
\par For all these algorithms, we choose all of the output feature bands to calculate the change intensity.

\subsection{Experiments on Taizhou ETM dataset}
\par
\begin{figure*}[htbp]
	\centering
	\includegraphics[scale=1.0]{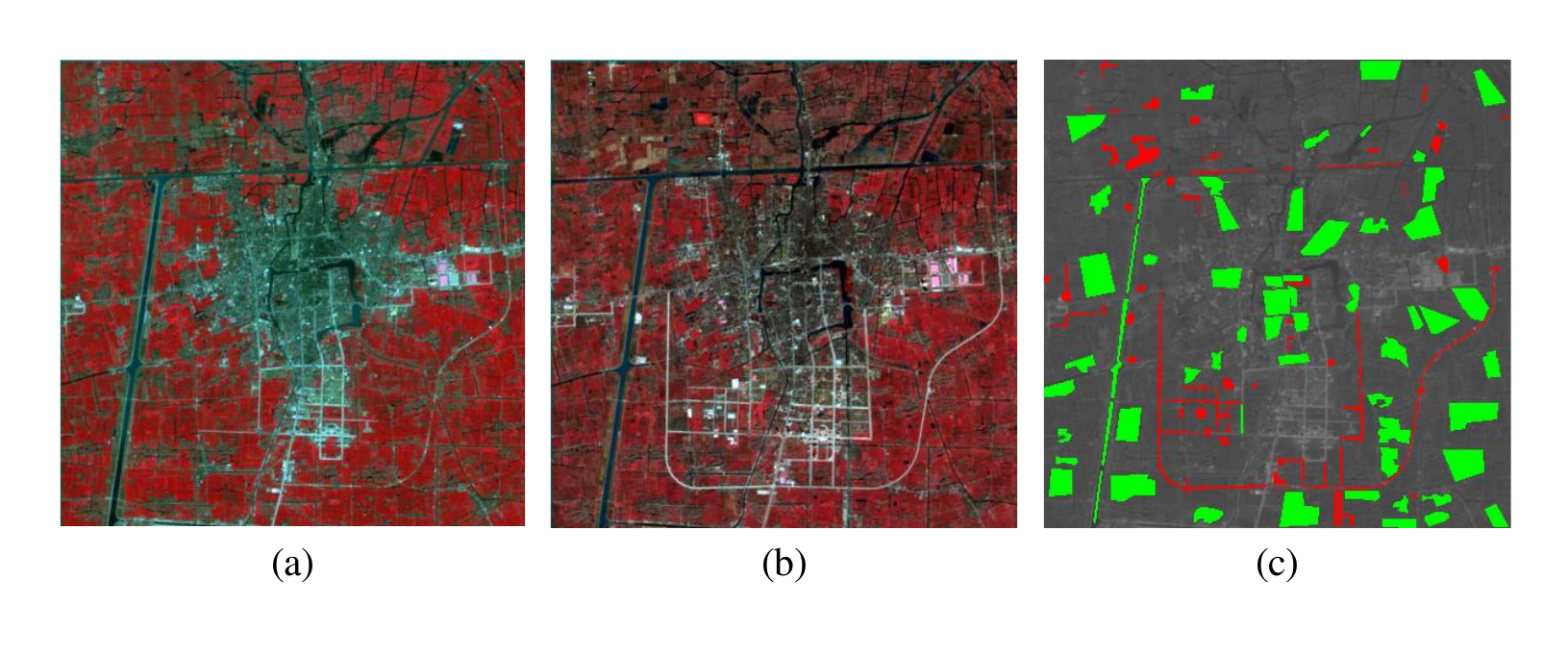}
	\caption{The pseudo-color images of Taizhou obtained in (a) 2000, (b) 2003, and (c) ground truth.}
	\label{taizhou}
\end{figure*}

 The study area of the first dataset is Taizhou city, Jiangsu Province, China. The image size is $400\times400$. Figure~\ref{taizhou} shows the pseudo color and ground truth images of this dataset. (a) and (b) are the pseudo color images acquired in 2002 and 2003, respectively. And (c) is the sampled ground truth image of changed and unchanged regions of Taizhou city, where the green pixels represent the unchanged regions, red pixels represent changed regions.{The background of image (c) is the gray scale image of (a), and they denote the unsampled regions.} The changed area contains 4227 pixels, and unchanged area contains 17163 pixels.

\par In the experiment of DSFA on Taizhou dataset, 4000 pixels, which are about 2.5\% of the total number of pixels, are randomly selected from the unchanged region of CVA pre-detected image for training to get the parameters of the networks and the projecting matrix of SFA. Due to the use of random initialization, for DSFA, we take the sum of change intensity of 10 independent runs as the final change intensity map, and the presented values of evaluation criteria are the results of the summed intensity map.

\par
\begin{figure*}[htbp]
	\centering
	\includegraphics[scale=0.8]{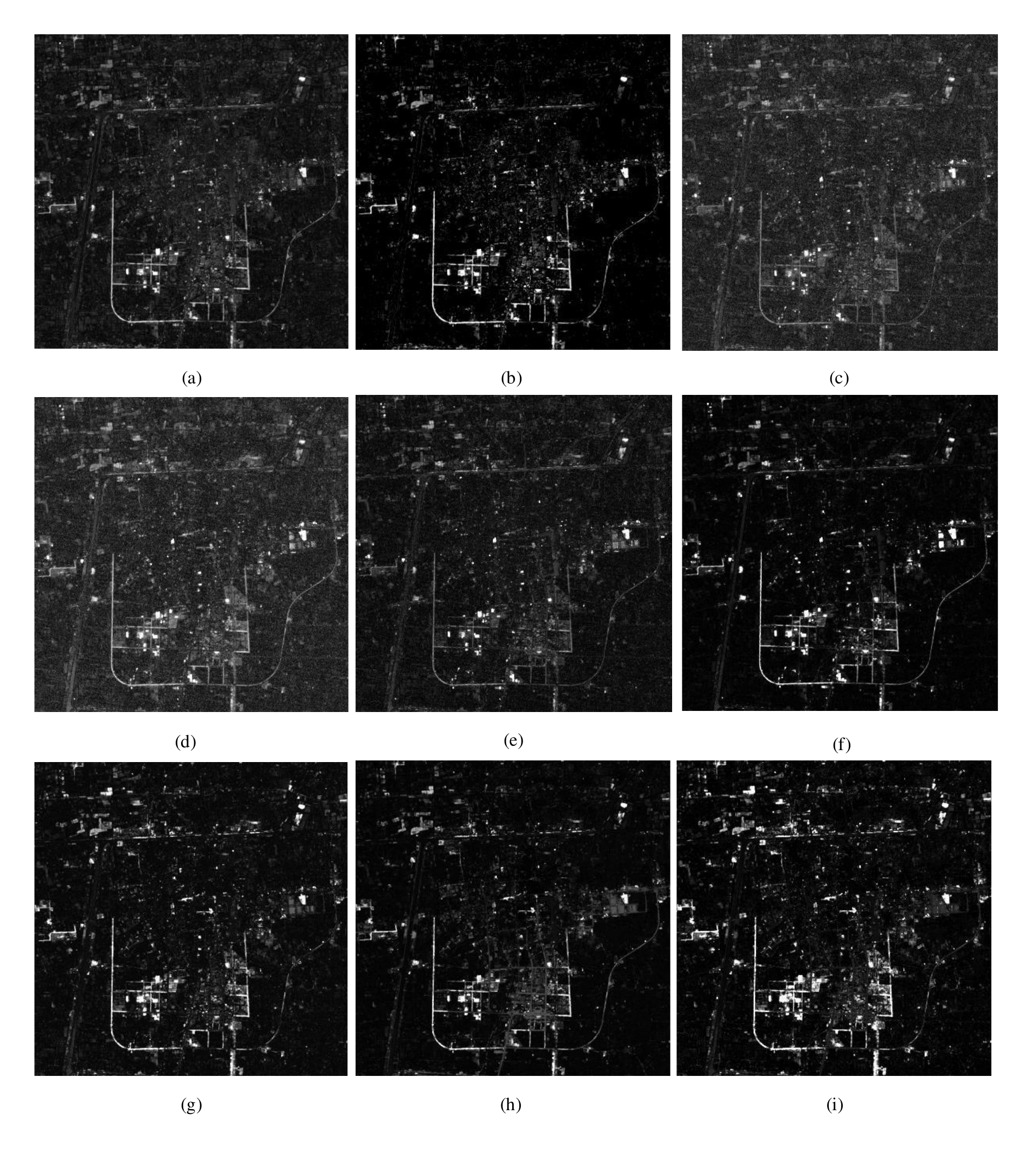}
	\caption{Change intensity maps of Taizhou dataset by (a) CVA, (b) PCA, (c) MAD, (d) IRMAD, (e) USFA, (f) ISFA, (g) DSFA-64-2, (h) DSFA-128-2, and (i) DSFA-256-2.}
	\label{taizhou_intensity}
\end{figure*}

 Figure~\ref{taizhou_intensity} shows the change intensity maps of Taizhou dataset by (a) CVA, (b) PCA, (c) MAD, (d) IRMAD, (e) USFA, (f) ISFA, (g) DSFA-64-2, (h) DSFA-128-2, and (i) DSFA-256-2. {Since PCANet and SDPCANet are both classification-based methods, there're no intensity maps of them.} DSFA-$h$-$l$ refers to a DSFA model with $l$ hidden layers and each hidden layer has $h$ nodes. All of these change intensity maps are calculated with all the output feature bands. In this figure, brighter regions have bigger change probabilities. As the Figure~\ref{taizhou_intensity} shows, visually, PCA, ISFA and DSFA-128-2 have the best performance in differentiating the changed and unchanged pixels. The unchanged regions of MAD and IRMAD are grey, which means they could not suppress the unchanged background from changed pixels very well. Similarly, CVA and USFA have bad performance in extracting changed pixels from unchanged background. As for other DSFA-based methods, DSFA-64-2 and DSFA-256-2, they have a moderate performance in change intensity map among all these methods. Though DSFA-based methods visually have some noise points, actually, these noise points probably represent truly changed pixels of unsampled region.

\par In Table~\ref{taizhou_otsu}, we present the accurate evaluation of binary change results segmented by OTSU method. {PCANet and SDPCANet are both classification-based methods, so their results presented here are their classification results and needn't to be processed by OTSU.} The evaluation criteria include the overall accuracy of sampled changed area (OA\_CHG), the overall accuracy of sampled unchanged area (OA\_UN), the overall accuracy of all sampled regions (OA), Kappa coefficient, and F1 score. The best values of each evaluation criteria are highlighted with bold.

\begin{table}[htbp]
	\centering
	\renewcommand{\arraystretch}{1.4}
	\caption{Change detection results of Taizhou dataset using OTSU.}
	\label{taizhou_otsu}
	\begin{tabular}{cccccc}
		\toprule[0.8pt]
		\textbf{OTSU}       & \textbf{OA\_CHG} & \textbf{OA\_UN} & \textbf{OA}     & \textbf{Kappa}  & \textbf{F1}     \\ \hline \hline
		\textbf{CVA}        & 0.8439           & 0.9970          & 0.9667          & 0.8890          & 0.9093          \\
		\textbf{PCA}        & 0.7755           & 0.9961          & 0.9525          & 0.8374          & 0.8658          \\ \hline
		\textbf{MAD}        & 0.8855           & 0.9474          & 0.9352          & 0.8030          & 0.8148          \\
		\textbf{IRMAD}      & 0.9056           & 0.9818          & 0.9667          & 0.8942          & 0.9150          \\ \hline
		\textbf{USFA}       & 0.7093           & 0.9922          & 0.9363          & 0.7773          & 0.8148          \\
		\textbf{ISFA}       & 0.8077           & 0.9991          & 0.9612          & 0.8684          & 0.8918          \\ \hline
		\textbf{PCANet}     & 0.8469           & \textbf{0.9992} & 0.9691          & 0.8967          & 0.9155          \\
		\textbf{SDPCANet}   & \textbf{0.9151}  & 0.9863          & 0.9722          & 0.9115          & 0.9287          \\ \hline
		\textbf{DSFA-64-2}  & 0.8294           & 0.9982          & 0.9648          & 0.8819          & 0.9032          \\
		\textbf{DSFA-128-2} & 0.8985           & 0.9954          & \textbf{0.9763} & \textbf{0.9227} & \textbf{0.9372} \\
		\textbf{DSFA-256-2} & 0.8450           & 0.9966          & 0.9667          & 0.8888          & 0.9090          \\ \hline
	\end{tabular}
\end{table}

\par As the Table~\ref{taizhou_otsu} shows, SDPCANet and PCANet have the best performance on OA\_CHG and OA\_UN, respectively. On detecting unchanged pixels, IRMAD has the second worst performance. On the contrary, ISFA performs bad on detection changed regions. And it is worth noting that DSFA-128-2 outperforms the other algorithms on OA, which indicates that it has a higher accuracy in both changed and unchanged part of remote sensing images. And other DSFA-based methods also have very good performance on OA, especially compared with USFA and ISFA. Besides, on Kappa coefficient, all DSFA-based methods have better performance than USFA and ISFA. The Kappa coefficient and F1 score of DSFA-128-2 are respectively 0.9227 and 0.9372, which are also much better than the other change detection methods. Considering the total detection accuracy of all changed and unchanged pixels, Kappa coefficient, and F1 score, DSFA-128-2 is the best method, {and SDPCANet is the second best method and only slightly worse than DSFA-128-2.}
\par The change detection results obtained by Kmeans method are presented in Table~\ref{taizhou_kmeans}. {PCANet and SDPCANet's results presented here also needn't to be processed by Kmeans.} As we can see from this table, all of these methods don't show obvious differences in performance when using different threshold algorithms. And this suggests that these methods, including our proposed DSFA-based algorithms, are robust to different threshold methods. The results in Table~\ref{taizhou_kmeans} are very similar to those in Table~\ref{taizhou_otsu}. {SDPCANet has the best performance on OA\_CHG, but shows lower accuracy on OA\_UN. On the contrary, PCANet is the best method in detecting unchanged regions, but has low accuracy in detecting changed pixels.} For both changed and unchanged regions, DSFA-128-2 has a detection accuracy of 97.64\%, which is still the highest among all methods. DSFA-128-2 also has the highest Kappa coefficient and F1 score. Generally, all of DSFA-based algorithms have pretty good performance. And among all the methods, DSFA-128-2 is still the best one.

\begin{table}[htbp]
	\centering
	\renewcommand{\arraystretch}{1.4}
	\caption{Change detection results of Taizhou dataset using Kmeans.}
	\label{taizhou_kmeans}
	\begin{tabular}{cccccc}
		\toprule[0.8pt]
		\textbf{Kmeans}     & \textbf{OA\_CHG} & \textbf{OA\_UN} & \textbf{OA}     & \textbf{Kappa}  & \textbf{F1}     \\ \hline\hline
		\textbf{CVA}        & 0.8453           & 0.9970          & 0.9670          & 0.8900          & 0.9102          \\
		\textbf{PCA}        & 0.7731           & 0.9964          & 0.9523          & 0.8365          & 0.8649          \\ \hline
		\textbf{MAD}        & 0.8827           & 0.9500          & 0.9367          & 0.8066          & 0.8464          \\
		\textbf{IRMAD}      & 0.9054           & 0.9818          & 0.9667          & 0.8942          & 0.9149          \\ \hline
		\textbf{USFA}       & 0.7166           & 0.9915          & 0.9372          & 0.7814          & 0.8185          \\
		\textbf{ISFA}       & 0.8074           & 0.9991          & 0.9612          & 0.8683          & 0.8916          \\ \hline
		\textbf{PCANet}     & 0.8469           & \textbf{0.9992} & 0.9691          & 0.8967          & 0.9155          \\
		\textbf{SDPCANet}   & \textbf{0.9151}  & 0.9863          & 0.9722          & 0.9115          & 0.9287          \\ \hline
		\textbf{DSFA-64-2}  & 0.8316           & 0.9981          & 0.9652          & 0.8830          & 0.9042          \\
		\textbf{DSFA-128-2} & 0.9006           & 0.9951          & \textbf{0.9764} & \textbf{0.9232} & \textbf{0.9377} \\
		\textbf{DSFA-256-2} & 0.8457           & 0.9966          & 0.9668          & 0.8892          & 0.9094          \\ \hline
	\end{tabular}
\end{table}

\par In Table~\ref{taizhou_best}, we present the best change detection results of Taizhou dataset by traversing of all thresholds. {Since SDPCANet ans PCANet don't need to be post-porocessed by threshold methods, their presented results are still based on their classification results.} In this table, we could see that all DSFA-based methods could outperform the other algorithms exclude CVA and ISFA. And among all DSFA-based methods, DSFA-128-2 has best performance in all evaluation criteria. ISFA has almost the same performance with DSFA-128-2. Besides, it's worth noting that the best change detection results of USFA and ISFA are much better than those obtained with OTSU and Kmeans method, while DSFA-based methods' best results are very close to those using OTSU and Kmeans method. We can conclude that though the best results of ISFA are very close to DSFA, the latter has much better discriminability than the former.

\begin{table}[htbp]
	\centering
	\renewcommand{\arraystretch}{1.4}
	\caption{ Best Change detection results of Taizhou dataset.}
	\label{taizhou_best}
	\begin{tabular}{cccccc}
		\toprule[0.8pt]
		\textbf{BEST}       & \textbf{OA}     & \textbf{Kappa}  & \textbf{F1}     \\ \hline\hline
		\textbf{CVA}        & 0.9756          & 0.9222          & 0.9373          \\
		\textbf{PCA}        & 0.9633          & 0.8810          & 0.9041          \\ \hline
		\textbf{MAD}        & 0.9472          & 0.8298          & 0.8626          \\
		\textbf{IRMAD}      & 0.9669          & 0.8945          & 0.9150          \\ \hline
		\textbf{USFA}       & 0.9476          & 0.8315          & 0.8640          \\
		\textbf{ISFA}       & 0.9776          & 0.9287          & 0.9426          \\ \hline
		\textbf{PCANet}     & 0.9691          & 0.8967          & 0.9155          \\
		\textbf{SDPCANet}   & 0.9722          & 0.9115          & 0.9287          \\ \hline
		\textbf{DSFA-64-2}  & 0.9715          & 0.9070          & 0.9254          \\
		\textbf{DSFA-128-2} & \textbf{0.9783} & \textbf{0.9304} & \textbf{0.9439} \\
		\textbf{DSFA-256-2} & 0.9713          & 0.9072          & 0.9250          \\ \hline
	\end{tabular}
\end{table}

\begin{figure*}[htbp]

	\centering
	\includegraphics[scale=0.4]{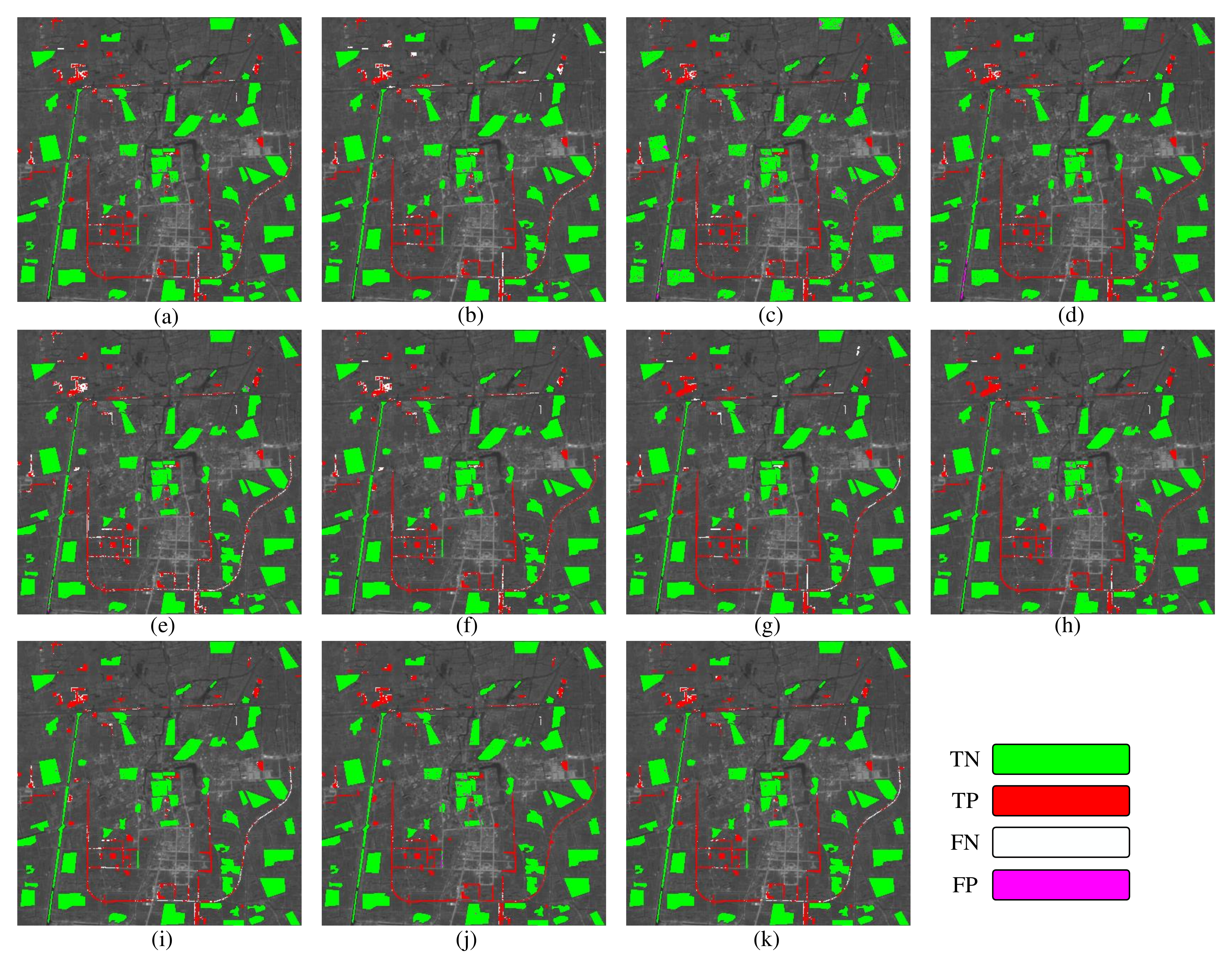}

	\caption{The binary change maps of (a) CVA, (b) PCA, (c) MAD, (d) IRMAD, (e) USFA, (f) ISFA, {(g) PCANet, (h) SDPCANet,} (i) DSFA-64-2, (j) DSFA-128-2 and (k) DSFA-256-2.}
	\label{taizhou_binary}
\end{figure*}

\par In Figure~\ref{taizhou_binary}, we present the binary change maps obtained by OTSU method of (a) CVA, (b) PCA, (c) MAD, (d) IRMAD, (e) USFA, (f) ISFA, (g) PCANet, (h) SDPCANet, (i) DSFA-64-2, (j) DSFA-128-2 and (k) DSFA-256-2. In this figure, green, red, white, and purple regions represent unchanged pixels that are detected as unchanged, changed pixels that are detected as changed, changed pixels that are detected as unchanged, and unchanged pixels that are detected as changed, respectively. And we could refer them as true negative, true positive, false negative, and false positive samples. As Figure~\ref{taizhou_binary} presents, intuitively, DSFA-128-2 have the best performance. And compared with DSFA-128-2, the results of MAD-based methods have more false positive pixels than other algorithms. CVA, PCA and two SFA-based methods tend to classify changed pixels as unchanged. {Compared with DSFA-128-2, PCANet has more false negative regions and SDPCANet has more false positive regions.} The other DSFA-based methods, DSFA-64-2 and DSFA-256-2, are prone to judge some specific changed regions as unchanged.

\subsection{Experiments on Nanjing ETM dataset}

\par
\begin{figure*}[htbp]
	\centering
	\includegraphics[scale=0.95]{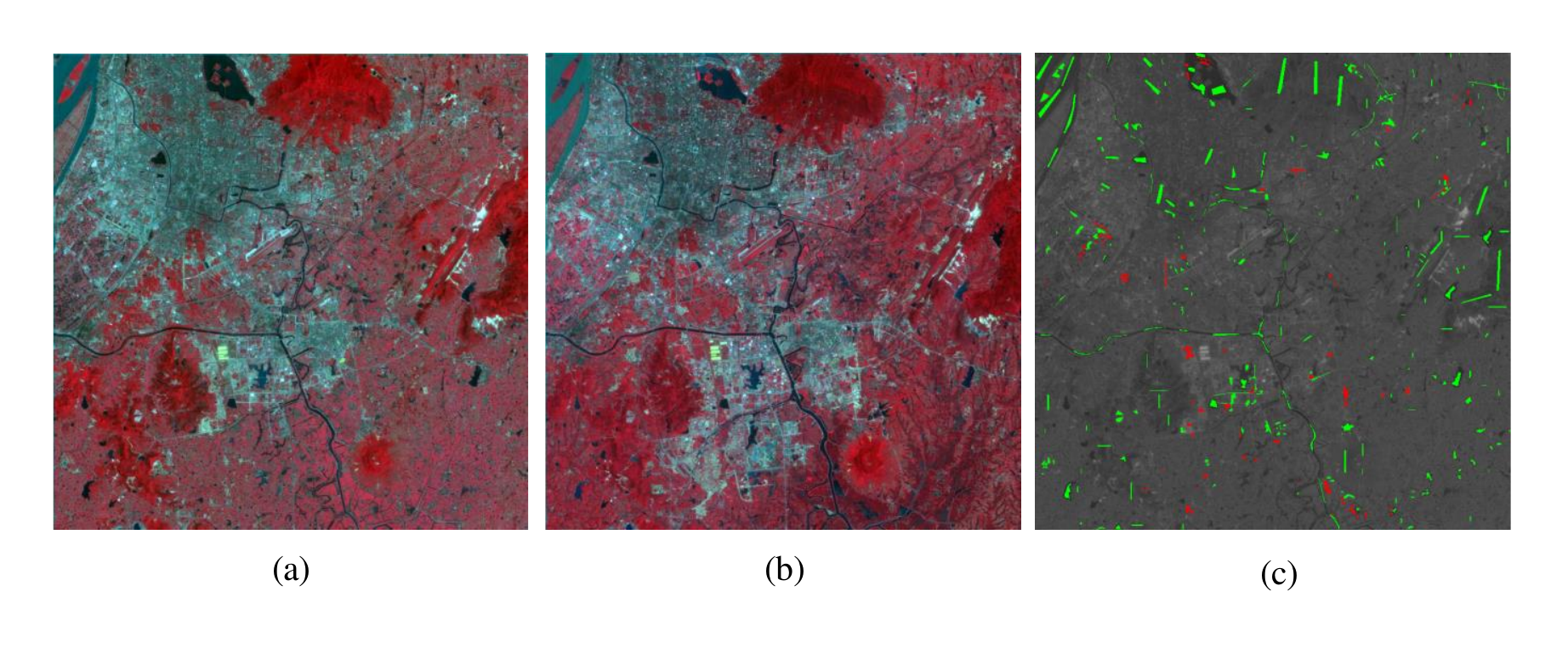}
	\caption{The pseudo-color images of Nanjing city obtained in (a) 2000, (b) 2002, and (c) ground truth.}
	\label{nanjing}
\end{figure*}

 The second experiment is carried on the Nanjing ETM dataset. Nanjing dataset includes two 6 spectral bands remote sensing images with a size of 800$\times$800, which are acquired in 2000 and 2002, respectively. Figure~\ref{nanjing} presents the pseudo color images of Nanjing city obtained in (a) 2000, (b) 2002, and (c) is the ground truth of sampled changed and unchanged areas. The red part of (c) represents the sampled changed area of Nanjing city, which includes 2363 pixels. And the green part is the sampled unchanged area and includes 12393 pixels.

\par In the experiment on Nanjing dataset, we randomly select 8000 pixels from unchanged area pre-detected by CVA to train our DSFA model. Like the experiment on Taizhou dataset, the presented results of each evaluation criteria of DSFA are based on the total change intensity map of 10 runs.
\par
\begin{figure*}[htbp]
	\centering
	\includegraphics[scale=0.52]{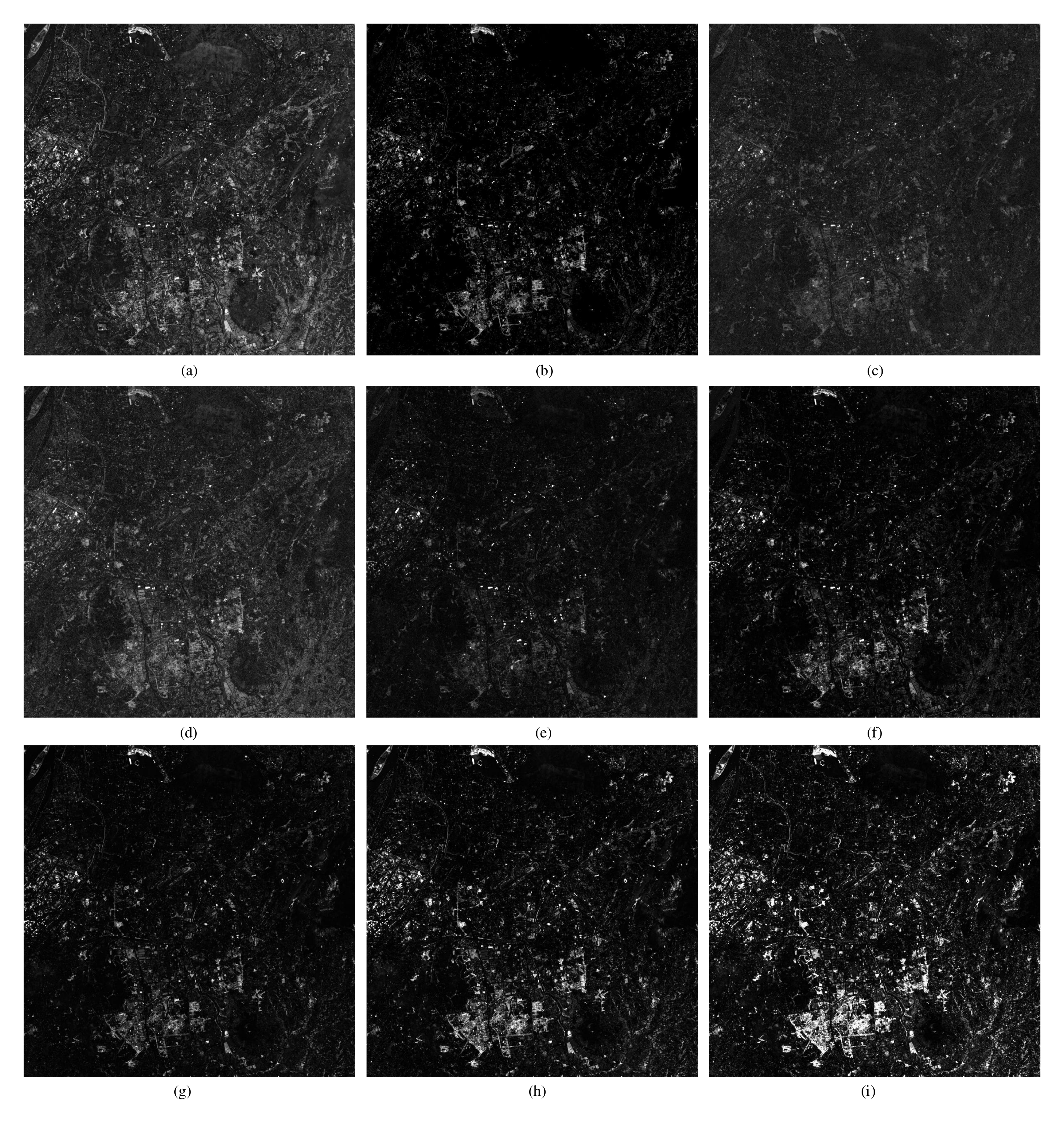}
	\caption{Change intensity maps of Nanjing dataset by (a) CVA, (b) PCA, (c) MAD, (d) IRMAD, (e) USFA, (f) ISFA, (g) DSFA-64-2, (h) DSFA-128-2, and (i) DSFA-256-2.}
	\label{nanjing_intensity}
\end{figure*}

 Figure~\ref{nanjing_intensity} shows the change intensity maps of Nanjing dataset by (a) CVA, (b) PCA, (c) MAD, (d) IRMAD, (e) USFA, (f) ISFA, (g) DSFA-64-2, (h) DSFA-128-2, and (i) DSFA-256-2. In this figure, brighter regions have bigger change probabilities. As we can see from this figure, USFA and ISFA have less bright area, which means that they tend to detect much less changed pixels that other change detection algorithms. And CVA, MAD and IRMAD have more bright area which indicates that thses methods are prone to categorize these unchanged pixels to changed. DSFA-128-2 and DSFA-256-2 have very close results to each other. Both them have very good discriminability of changed and unchanged pixels. In addition, the result of PCA is also very close to DSFA-64-2. But the distinction between their changed and unchanged regions is not very obvious. On the whole, visually, the result of DSFA-128-2 is the best in calculating the change intensity.

\par In Table~\ref{nanjing_otsu}, we present the change detection results of Nanjing dataset utilizing OTSU method. The best values of each evaluation criteria are highlighted with bold in this table. As we can see, in general, DSFA-based methods, especially DSFA-128-2, have the best performance among all these methods. DSFA-128-2 could outperform other algorithms on OA\_UN, OA, Kappa coefficient and F1 score. And in these criteria, all DSFA-based methods are much better than others. MAD and IRMAD have the best performance on OA\_CHG, which is consistent with their change intensity results. Similar to MAD and IRMAD, CVA and PCA have very high values on OA\_CHG, but are far worse than DSFA-based methods on OA\_UN, OA and Kappa coefficient. {The results of PCANet and SDPCANet are also very similar to the results of PCA method.} On the contrary, USFA and ISFA do well in detecting unchanged pixels, but have the lowest accuracy on OA\_CHG.

\begin{table}[htbp]
	\centering
	\renewcommand{\arraystretch}{1.4}
	\caption{Change detection results of Nanjing dataset using OTSU.}
	\label{nanjing_otsu}
	\begin{tabular}{cccccc}
		\toprule[0.8pt]
		\textbf{OTSU}       & \textbf{OA\_CHG} & \textbf{OA\_UN} & \textbf{OA}     & \textbf{Kappa}  & \textbf{F1}     \\ \hline\hline
		\textbf{CVA}        & 0.8595           & 0.9168          & 0.9076          & 0.6933          & 0.7487          \\
		\textbf{PCA}        & 0.8625           & 0.9363          & 0.9244          & 0.7398          & 0.7853          \\ \hline
		\textbf{MAD}        & \textbf{0.9534}  & 0.8530          & 0.8691          & 0.6236          & 0.6999          \\
		\textbf{IRMAD}      & 0.9530           & 0.8922          & 0.9019          & 0.6987          & 0.7568          \\ \hline
		\textbf{USFA}       & 0.5959           & 0.9680          & 0.9084          & 0.6234          & 0.6757          \\
		\textbf{ISFA}       & 0.6416           & 0.9760          & 0.9224          & 0.6816          & 0.7260          \\ \hline
		\textbf{PCANet}     & 0.8680           & 0.9334          & 0.9229          & 0.7367          & 0.7829          \\
		\textbf{SDPCANet}   & 0.8426           & 0.9397          & 0.9242          & 0.7351          & 0.7806          \\ \hline
		\textbf{DSFA-64-2}  & 0.7288           & \textbf{0.9817} & 0.9412          & 0.7647          & 0.7987          \\
		\textbf{DSFA-128-2} & 0.7465           & 0.9806          & \textbf{0.9431} & \textbf{0.7747} & \textbf{0.8078} \\
		\textbf{DSFA-256-2} & 0.7360           & 0.9793          & 0.9403          & 0.7633          & 0.7980          \\ \hline
	\end{tabular}
\end{table}

\par Table~\ref{nanjing_kmeans} shows the evaluation results of the experiment on Nanjing dataset using Kmeans method. Similar to the results of OTSU, compared to MAD-based and SFA-based methods, DSFA is still better in detecting unchanged and changed areas, respectively. On the whole, DSFA-based algorithms have higher overall accuracies, Kappa values and F1 score than others. {PCANet-based methods have higher OA\_CHG than DSFA-based methods, but worse performances on the other criteria. In general, PCANet-based methods is the second best.}

\begin{table}[htbp]
	\centering
	\renewcommand{\arraystretch}{1.4}
	\caption{Change detection results of Nanjing dataset using Kmeans.}
	\label{nanjing_kmeans}
	\begin{tabular}{cccccc}
		\toprule[0.8pt]
		\textbf{Kmeans}     & \textbf{OA\_CHG} & \textbf{OA\_UN} & \textbf{OA}     & \textbf{Kappa}  & \textbf{F1}     \\ \hline\hline
		\textbf{CVA}        & 0.8578           & 0.9184          & 0.9087          & 0.6958          & 0.7506          \\
		\textbf{PCA}        & 0.8650           & 0.9352          & 0.9240          & 0.7390          & 0.7846          \\ \hline
		\textbf{MAD}        & 0.9518           & 0.8557          & 0.8711          & 0.6276          & 0.7028          \\
		\textbf{IRMAD}      & \textbf{0.9564}  & 0.8882          & 0.8991          & 0.6924          & 0.7523          \\ \hline
		\textbf{USFA}       & 0.5832           & 0.9692          & 0.9074          & 0.6159          & 0.6685          \\
		\textbf{ISFA}       & 0.6437           & 0.9760          & 0.9227          & 0.6833          & 0.7275          \\ \hline
		\textbf{PCANet}     & 0.8680           & 0.9334          & 0.9229          & 0.7367          & 0.7829          \\
		\textbf{SDPCANet}   & 0.8426           & 0.9397          & 0.9242          & 0.7351          & 0.7806          \\ \hline
		\textbf{DSFA-64-2}  & 0.7290           & \textbf{0.9817} & 0.9412          & 0.7647          & 0.7987          \\
		\textbf{DSFA-128-2} & 0.7463           & 0.9807          & \textbf{0.9432} & \textbf{0.7748} & \textbf{0.8079} \\
		\textbf{DSFA-256-2} & 0.7361           & 0.9792          & 0.9403          & 0.7632          & 0.7980          \\ \hline
	\end{tabular}
\end{table}

\par In Table~\ref{nanjing_best}, we present the best threshold result of each changed detection methods by traversing all values. We could see from this table that DSFA-based methods are still the best on all the criteria. PCA, IRMAD and ISFA have high values on F1 score, but are much worse on OA and Kappa than DSFA-based methods. Besides, it's also worth noting that the best results of DSFA-based methods are very close to the results obtained by OTSU and Kmeans, which could be an evidence of the good discriminability of DSFA's results. On the contrary, threshold results and the best results of USFA and ISFA have a sensible difference. And the best results of CVA, PCA and MAD-based methods are also much better than their threshold results in both OA and Kappa coefficient.

\begin{table}[htbp]
	\centering
	\renewcommand{\arraystretch}{1.4}
	\caption{ Best Change detection results of Nanjing dataset.}
	\label{nanjing_best}

	\begin{tabular}{cccccc}
		\toprule[0.8pt]
		\textbf{BEST}       & \textbf{OA}     & \textbf{Kappa}  & \textbf{F1}     \\ \hline\hline
		\textbf{CVA}        & 0.9248          & 0.7178          & 0.7652          \\
		\textbf{PCA}        & 0.9341          & 0.7518          & 0.7925          \\ \hline
		\textbf{MAD}        & 0.9227          & 0.7244          & 0.7725          \\
		\textbf{IRMAD}      & 0.9229          & 0.7340          & 0.7815          \\ \hline
		\textbf{USFA}       & 0.9164          & 0.6997          & 0.7517          \\
		\textbf{ISFA}       & 0.9336          & 0.7578          & 0.7984          \\ \hline
		\textbf{PCANet}     & 0.9229          & 0.7367          & 0.7829          \\
		\textbf{SDPCANet}   & 0.9242          & 0.7351          & 0.7806          \\ \hline
		\textbf{DSFA-64-2}  & \textbf{0.9450} & \textbf{0.7915} & \textbf{0.8244} \\
		\textbf{DSFA-128-2} & 0.9439          & 0.7850          & 0.8195          \\
		\textbf{DSFA-256-2} & 0.9409          & 0.7664          & 0.8015          \\ \hline
	\end{tabular}
\end{table}

\begin{figure*}[htbp]

	\centering
	\includegraphics[scale=0.215]{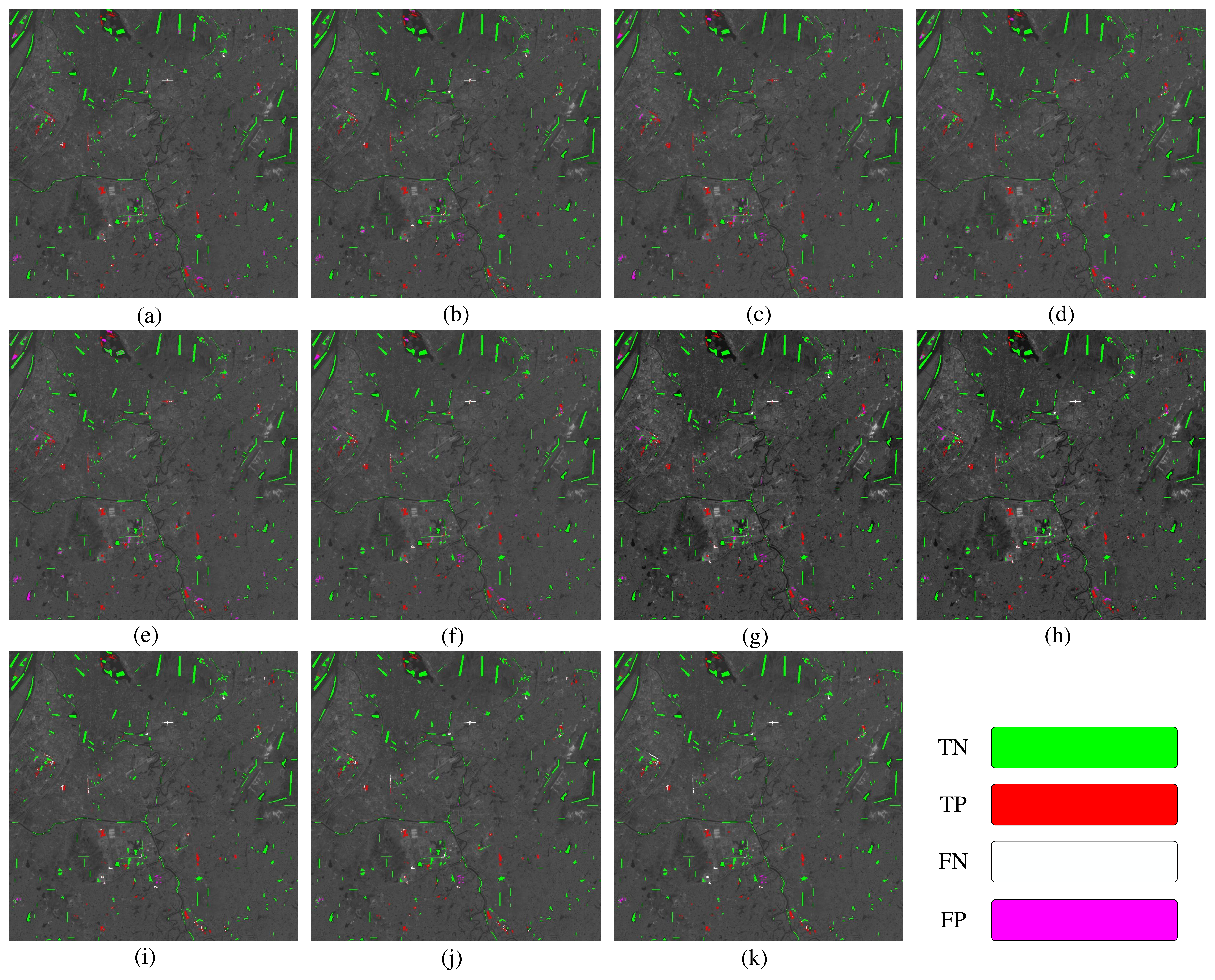}
	\caption{The binary change maps of (a) CVA, (b) PCA, (c) MAD, (d) IRMAD, (e) USFA, (f) ISFA, (g) PCANet, (h) SDPCANet, (i) DSFA-64-2, (j) DSFA-128-2 and (k) DSFA-256-2.}
	\label{nanjing_binary}
\end{figure*}

\par Figure~\ref{nanjing_binary} shows the binary change maps of (a) CVA, (b) PCA, (c) MAD, (d) IRMAD, (e) USFA, (f) ISFA, (g) PCANet, (h) SDPCANet, (i) DSFA-64-2, (j) DSFA-128-2 and (k) DSFA-256-2, which are segmented by OTSU method. According to this figure, we could see that the binary change result of DSFA with different net structure are almost the same. Obviously, compared with DSFA's results, results of MAD and IRMAD have much more purple pixels, which represent the false positive samples. On the contrary, results of USFA and ISFA contain more false negative pixels, which are colored with white. The results of CVA and PCA are close to DSFA's results, but still has less true negative and more false positive samples than the latter. {Besides, PCANet and SDPCANet also have a higher false positive rate than DSFA-based methods}.

{

\subsection{Experiments on River dataset}
\par The River dataset consists of two 198 bands images with a spatial size of $463\times241$. The changed regions of this dataset contain 12566 pixels, while the unchanged regions contain 99017 pixels. Figure~\ref{river} presents the bi-temporal images and ground truth map of River dataset. In Figure~\ref{river}, changed regions are white and unchanged regions are black.

\begin{figure}[htbp]
	\centering
	{
		\includegraphics[scale=0.4]{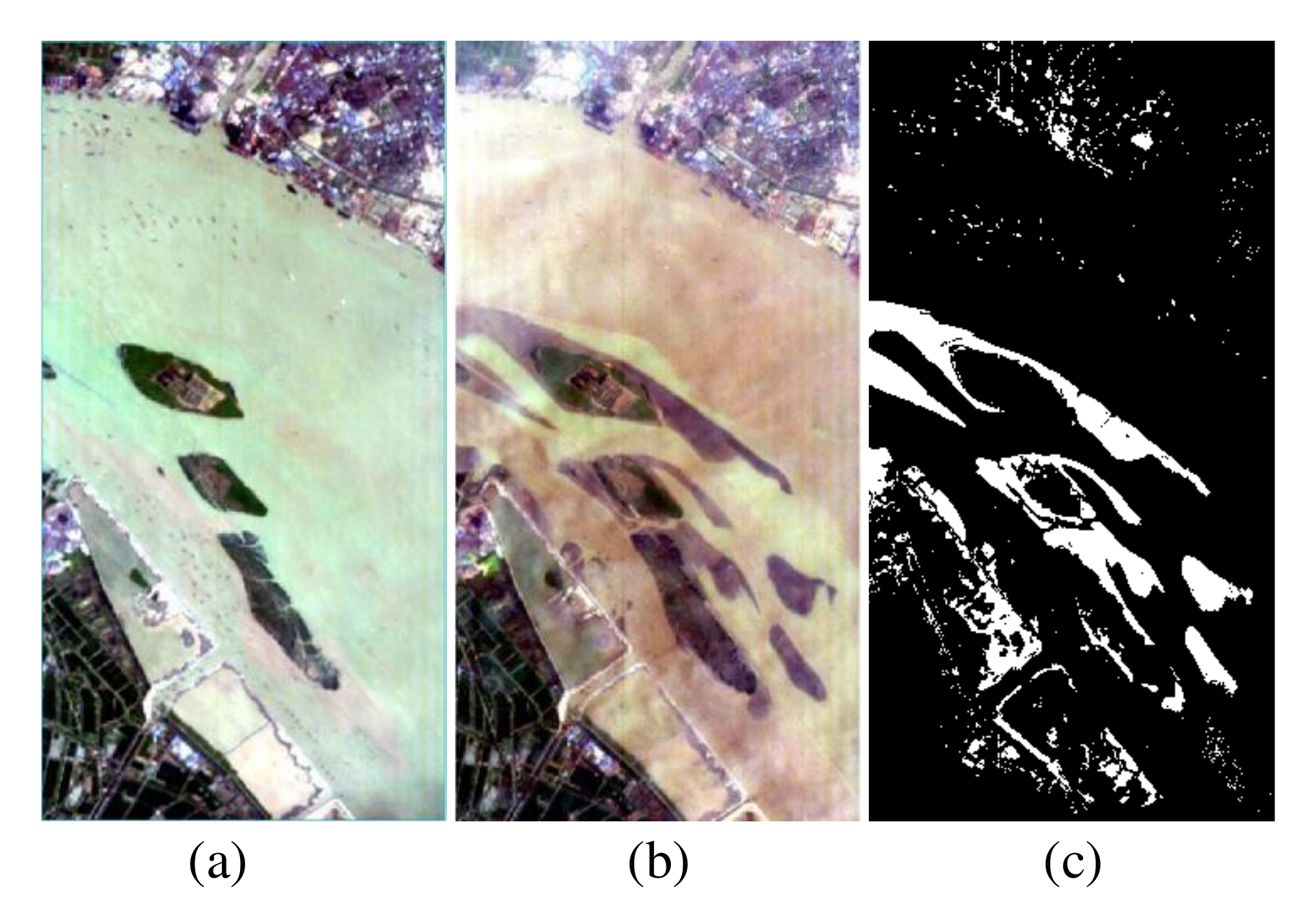}
		\caption{The bi-temporal imagery of River dataset obtained in (a) May, 2013, (b) Dec, 2013, and (c) ground truth map.}
		\label{river}}
\end{figure}
\par In Figure~\ref{river_intensity}, we present the change intensity maps of our proposed methods and all control methods. PCANet and SDPCANet are based on classification, so there's no relevant intensity maps in this figure. As can be observed from Figure~\ref{river_intensity}, intuitively, all DSFA-based methods have better discriminability than CVA, PCA, and methods based on MAD and USFA. CVA, PCA and ISFA also have a better performance in separating the changed and unchanged regions than MAD, IRMAD and USFA. Visually, compared to the ground truth map, DSFA-based methods have relatively high false negative rate in the upper-right area of imagery. And other methods have brighter upper-right and lower-left area, which suggests that these methods are prone to detect these areas as changed, while most of them are unchanged actually.

\begin{figure*}[htbp]
	\centering
	{
		\includegraphics[scale=0.30]{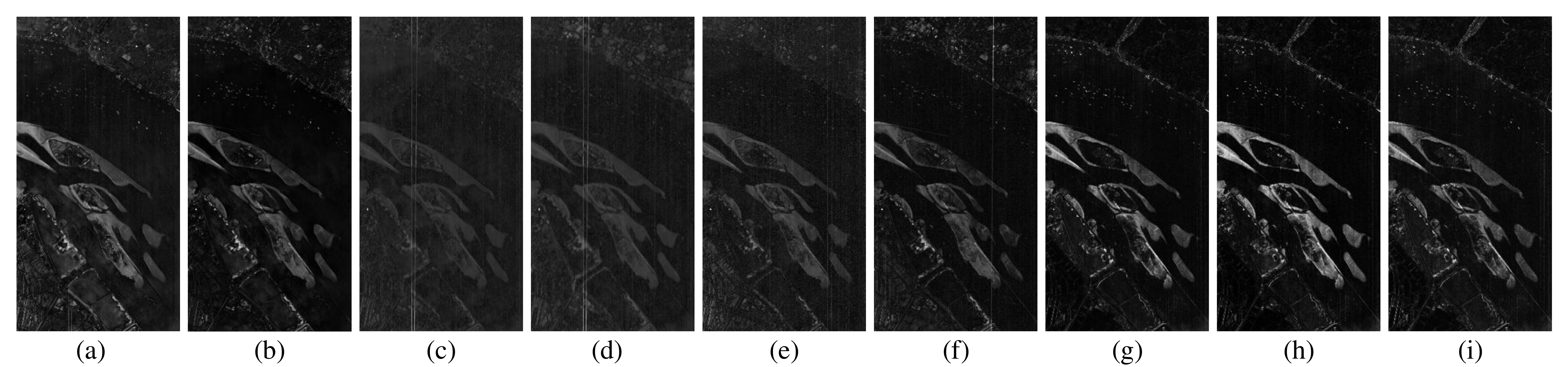}
		\caption{The change intensity maps of (a) CVA, (b) PCA, (c) MAD, (d) IRMAD, (e) USFA, (f) ISFA, (g) DSFA-64-2, (h) DSFA-128-2, and (i) DSFA-256-2.}
		\label{river_intensity}}
\end{figure*}

\par We then use OTSU and Kmeans method to obtain the results with different criteria using the aforementioned methods. The obtained numerical results, along with the results of PCANet and SDPCANet, are presented in Table~\ref{river_otsu_kmeans}. The best value of each column is highlighted in bold in this table.
\par As could be observed from Table~\ref{river_otsu_kmeans}, DSFA-based method could achieve better performance on OA\_UN, OA, Kappa and F1 score. Among all these methods, DSFA-128-2 has the best performance on OA, Kappa and F1 score, and the third best performance on OA\_UN. DSFA-64-2 and SDPCANet both have the highest accuracy on OA\_UN. Though PCANet have high performance on OA\_CHG and F1 score, its performance on OA\_UN, OA and Kappa are much worse than DSFA-based methods. In addition, it's also worth noting that the results using Kmeans and OTSU of our proposed methods still show very slight differences, which indicates that our proposed DSFA method are robust to different threshold methods.

\begin{table*}[htbp]
	\centering

	{

		\renewcommand{\arraystretch}{1.4}
		\caption{Change detection results of River dataset.}
		\label{river_otsu_kmeans}

		\begin{tabular}{ccccccccccc}
			\toprule[0.8pt]
			& \multicolumn{5}{c}{\textbf{Kmeans}}                                                                           & \multicolumn{5}{c}{\textbf{OTSU}}                                                        \\ \cline{2-11}
			                    & \textbf{OA\_CHG} & \textbf{OA\_UN} & \textbf{OA}     & \textbf{Kappa}  & \textbf{F1}                          & \textbf{OA\_CHG} & \textbf{OA\_UN} & \textbf{OA}     & \textbf{Kappa}  & \textbf{F1}     \\ \hline  \hline
			\textbf{CVA}        & 0.8168           & 0.9082          & 0.8979          & 0.5868          & \multicolumn{1}{c|}{0.6432}          & \textbf{0.8712}  & 0.8770          & 0.8764          & 0.5474          & 0.6135          \\
			\textbf{PCA}        & 0.5899           & 0.9532          & 0.9123          & 0.5531          & \multicolumn{1}{c|}{0.6024}          & 0.5734           & 0.9560          & 0.9129          & 0.5484          & 0.5971          \\ \hline
			\textbf{MAD}        & 0.8022           & 0.9142          & 0.9016          & 0.5927          & \multicolumn{1}{c|}{0.6474}          & 0.8563           & 0.8864          & 0.8830          & 0.5591          & 0.6223          \\
			\textbf{IRMAD}      & 0.8093           & 0.9130          & 0.9013          & 0.5940          & \multicolumn{1}{c|}{0.6488}          & 0.8271           & 0.9059          & 0.8970          & 0.5872          & 0.6440          \\ \hline
			\textbf{USFA}       & \textbf{0.8297}  & 0.8953          & 0.8879          & 0.5638          & \multicolumn{1}{c|}{0.6250}          & 0.8400           & 0.8871          & 0.8818          & 0.5514          & 0.6155          \\
			\textbf{ISFA}       & 0.6127           & 0.9377          & 0.9011          & 0.5267          & \multicolumn{1}{c|}{0.5826}          & 0.6377           & 0.9314          & 0.8984          & 0.5281          & 0.5856          \\ \hline
			\textbf{PCANet}     & 0.8024           & 0.9487          & 0.9322          & 0.6889          & \multicolumn{1}{c|}{0.7273}          & 0.8024           & 0.9487          & 0.9322          & 0.6889          & 0.7273          \\
			\textbf{SDPCANet}   & 0.5393           & \textbf{0.9850} & 0.9348          & 0.6166          & \multicolumn{1}{c|}{0.6507}          & 0.5393           & 0.9850          & 0.9348          & 0.6166          & 0.6507          \\ \hline
			\textbf{DSFA-64-2}  & 0.6164           & 0.9848          & 0.9434          & 0.6796          & \multicolumn{1}{c|}{0.7293}          & 0.6134           & \textbf{0.9851} & 0.9432          & 0.6780          & 0.7102          \\
			\textbf{DSFA-128-2} & 0.6877           & 0.9812          & \textbf{0.9482} & \textbf{0.7207} & \multicolumn{1}{c|}{\textbf{0.7508}} & 0.6864           & 0.9815          & \textbf{0.9483} & \textbf{0.7206} & \textbf{0.7494} \\
			\textbf{DSFA-256-2} & 0.6622           & 0.9777          & 0.9422          & 0.6888          & \multicolumn{1}{c|}{0.7283}          & 0.6615           & 0.9778          & 0.9422          & 0.6884          & 0.7207          \\ \hline
		\end{tabular}
	}
\end{table*}

\par The best results of each method are obtained by traversing all possible thresholds, and are presented in Table~\ref{river_best}. DSFA-based methods still have the best performance. Specifically, DSFA methods have much better performance on OA, Kappa and F1 score than other methods. Actually, DSFA-128-2 could outperform all other methods on all criteria. DSFA-64-2 and DSFA-256-2 respectively have the second and third best OA and Kappa value, and they're very close to DSFA-128-2 on F1 score. In addition, the best values of DSFA methods are only slightly better than the results obtained with threshold methods, which also suggests that the transformed features of DSFA have a better discriminability.
\begin{table}[htbp]
	\centering
	{
		\renewcommand{\arraystretch}{1.4}
		\caption{ Best Change detection results of River dataset.}
		\label{river_best}

		\begin{tabular}{cccccc}
			\toprule[0.8pt]
			\textbf{BEST}       & \textbf{OA}     & \textbf{Kappa}  & \textbf{F1}     \\ \hline\hline
			\textbf{CVA}        & 0.9264          & 0.6242          & 0.6841          \\
			\textbf{PCA}        & 0.9204          & 0.6075          & 0.6641          \\ \hline
			\textbf{MAD}        & 0.9140          & 0.5972          & 0.6481          \\
			\textbf{IRMAD}      & 0.9095          & 0.5984          & 0.6510          \\ \hline
			\textbf{USFA}       & 0.9180          & 0.6098          & 0.6590          \\
			\textbf{ISFA}       & 0.9098          & 0.5285          & 0.5879          \\ \hline
			\textbf{PCANet}     & 0.9322          & 0.6889          & 0.7273          \\
			\textbf{SDPCANet}   & 0.9348          & 0.6166          & 0.6507          \\ \hline
			\textbf{DSFA-64-2}  & 0.9454          & 0.7109          & 0.7419          \\
			\textbf{DSFA-128-2} & \textbf{0.9483} & \textbf{0.7270} & \textbf{0.7566} \\
			\textbf{DSFA-256-2} & 0.9423          & 0.7007          & 0.7344          \\ \hline
		\end{tabular}
	}
\end{table}
\par In Figure~\ref{river_binary}, the binary change maps obtained by different methods are presented. Consistent with the results in Figure~\ref{river_intensity}, DSFA algorithms have lower accuracies in detecting the changes in the upper-right regions of the original images, but have much better performance in other regions. The changes in the upper-right regions are not apparent and the background is complex, which we think is the main reason of DSFA's lower accuracy. On the contrary, CVA, PCA, MAD-based and SFA-based methods have a relatively high flase positive rate in both the upper-right regions and lower-left regions. It's also noticed that SDPCANet also has a high flase negative rate in the upper-right region, and PCANet tends to categorize the unchanged pixels in the lower-left regions as changed. On the whole, DSFA methods have the best performance visually and numerically.

\begin{figure*}[htbp]

	\centering
	\includegraphics[scale=0.239]{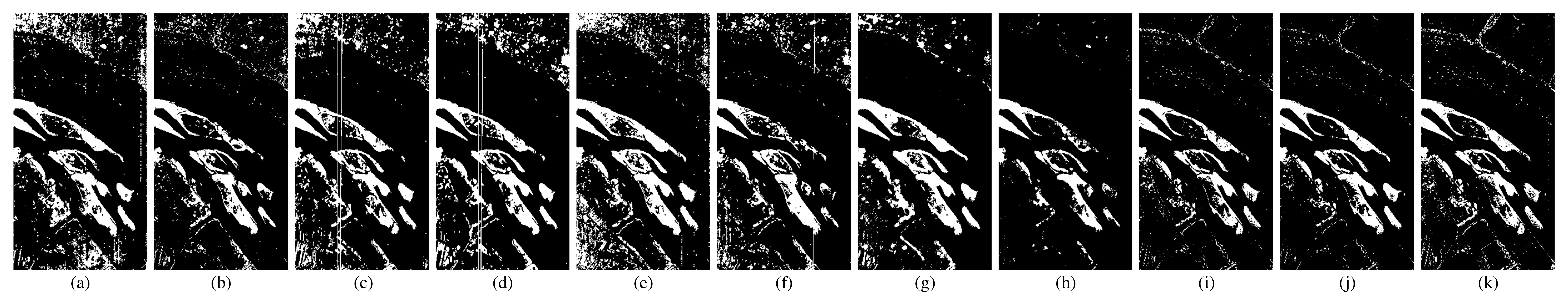}
	{
		\caption{The binary change maps of (a) CVA, (b) PCA, (c) MAD, (d) IRMAD, (e) USFA, (f) ISFA, (g) PCANet, (h) SDPCANet, (i) DSFA-64-2, (j) DSFA-128-2 and (k) DSFA-256-2.}
		\label{river_binary}}
\end{figure*}

\subsection{Runtime Analysis}
\par Though our proposed DSFA is based on fully connected networks, it's actually not very time consuming compared with other methods. We present the comparison of the runtime of IRMAD, ISFA, DSFA-128-2, PCANet and SDPCANet on three datasets in Figure~\ref{runtime}. IRMAD and ISFA are implemented with MATLAB and run on CPU. PCANet and SDPCANet are also implemented with MATLAB but accelerated with 12 threads. DSFA-128-2 is implemented with Python and runs on CPU and GPU separately, which are respectively denoted by DSFA-CPU and DSFA-GPU in Figure~\ref{runtime}. The CPU used is Intel Xeon E5 with a clock rate of 2.2 GHz. The GPU used is a single NVIDAI 1080Ti card.

\begin{figure}[htbp]
	\centering
	{
		\includegraphics[scale=0.50]{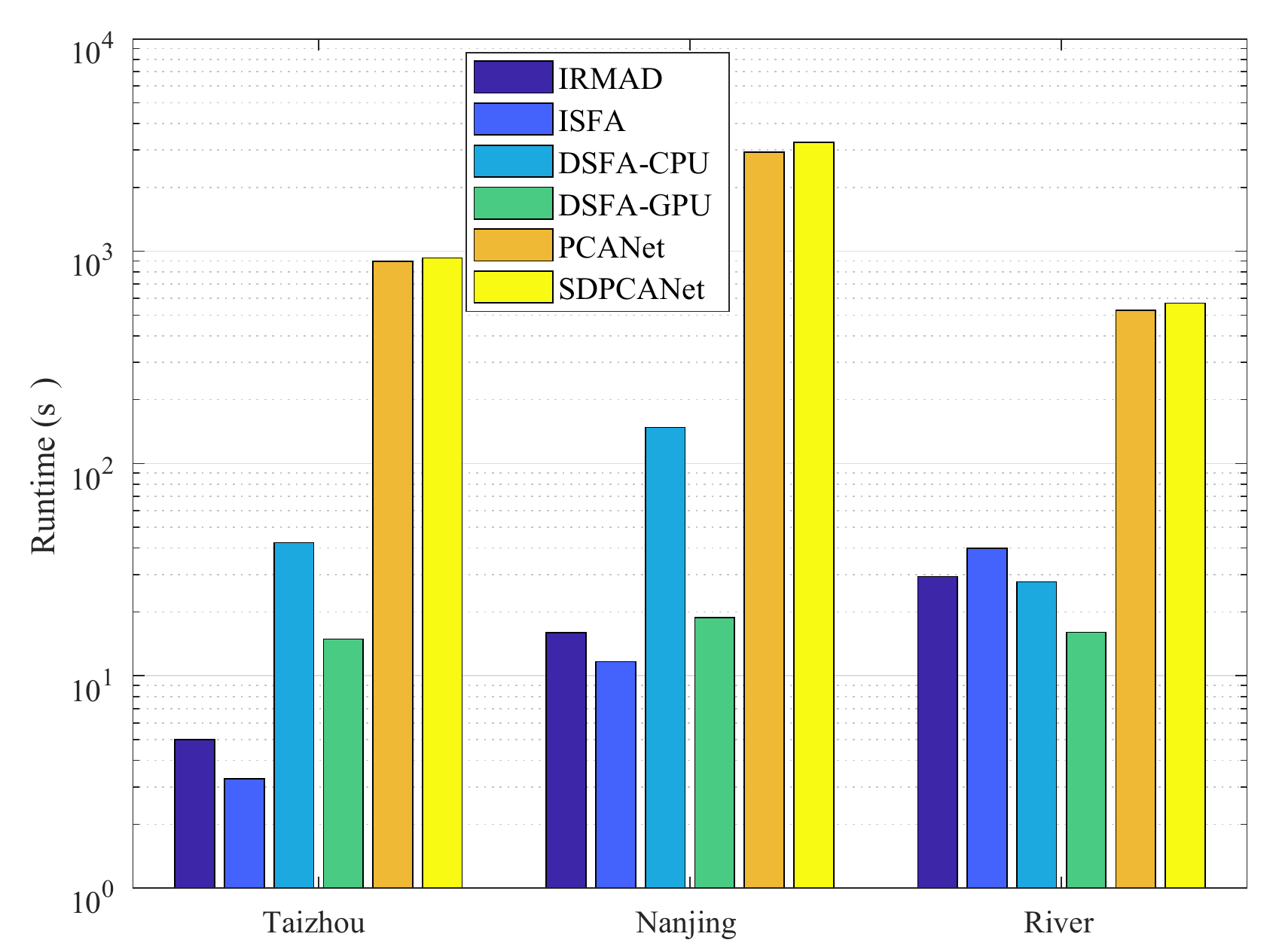}
		\caption{The comparison of runtime of different change detection methods.}
		\label{runtime}}
\end{figure}

\par As presents in this figure, ISFA and IRMAD are the two fastest methods, followed by DSFA-GPU and DSFA-CPU. Two PCANet-based method are the most time consuming. Besides, DSFA-GPU and DSFA-CPU are both faster than IRMAD and ISFA on River dataset, due to the smaller image size and more spectral bands of this dataset. On Taizhou and Nanjing dataset, the runtime of DSFA-GPU is very close to ISFA and IRMAD. DSFA-CPU is a little more time consuming, but it's still acceptable considering its improvements than IRMAD and ISFA.

\section{Discussion}\label{discussion}

\subsection{Hyperparameter Analysis}
\par In our experiments, we take $10^{-4}$ as the value of the regularization parameter $r$ in Equation~(22-23). However, in fact, $r$ does not have significant influence on the final results when it's small enough.
\par In Figure~\ref{r}, we present the relationship curves between the final change detection accuracy and $r$ on three datasets. The network used is DSFA-128-2. It can be observed that when $r<10^{-4}$, the accuracy curves on three dataset only have ignorable changes. On the contrary, when $r>10^{-4}$, the accuracies are much lower because a larger $r$ may corrupt the characteristic of the covariance matrices in Equation~(22-23).
\begin{figure}[htbp]
	\centering
	{
		\includegraphics[scale=0.45]{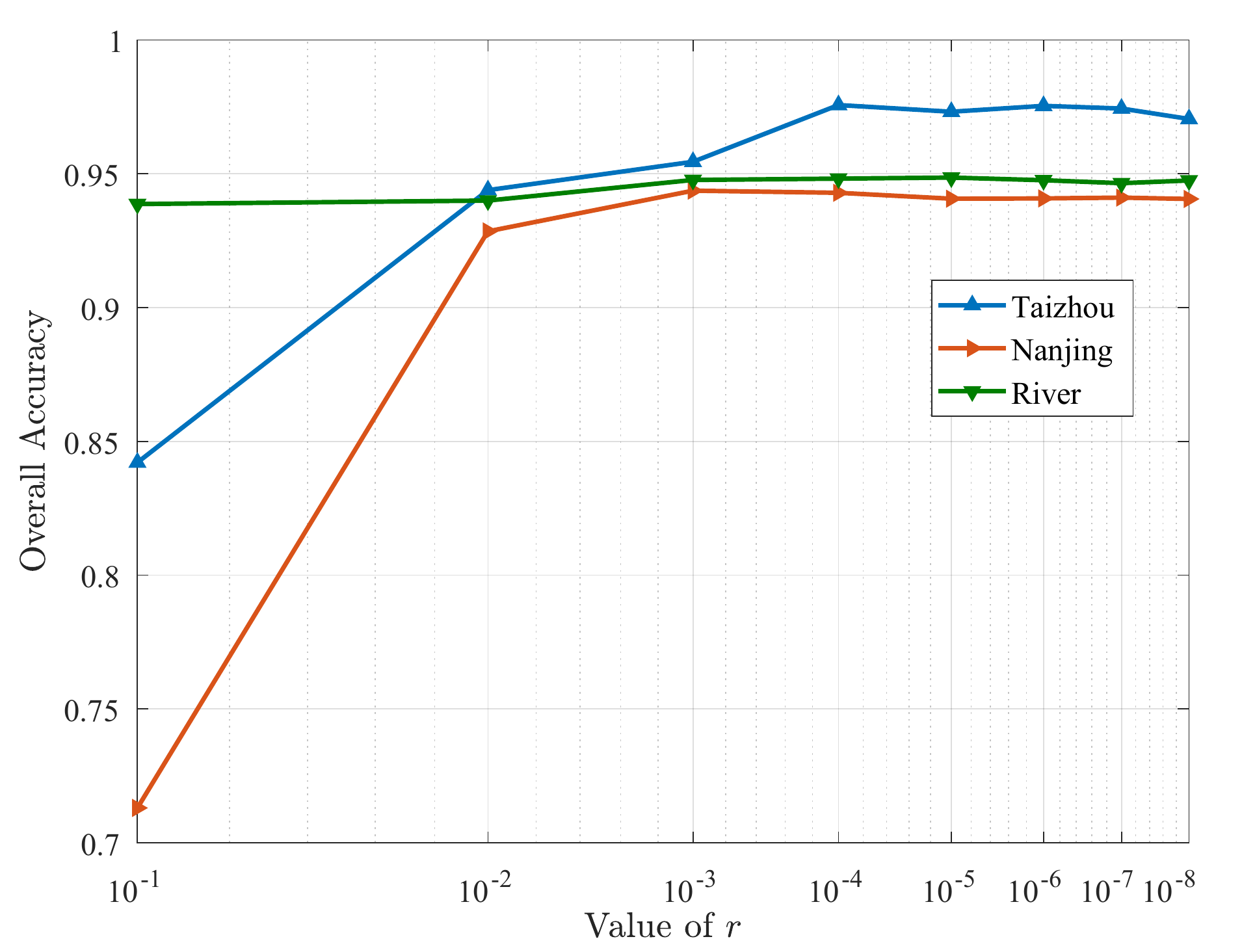}
		\caption{The relationship between $r$ and OA on three datasets.}
		\label{r}}
\end{figure}

\subsection{Selection of Training Samples}
\par In the Figure~\ref{selection}, we present the final accuracies using difference training sample selection strategies. This experiment is performed on the River dataset using DSFA-128-2. In Figure~\ref{selection}, Negative and Ground Truth strategy respectively mean that training samples are selected from the changed and unchanged regions of the ground truth image. Random strategy means training samples are absolutely randomly selected from the original imagery. And CVA strategy denotes that the training samples are selected from the unchanged regions of the change detection results of CVA.
\par As shown in this figure, Negative strategy leads to a very bad result, since the learned projection from changed pixel pairs conflicts with the main idea of SFA and DSFA. Random strategy is very slightly better than CVA and Ground Truth strategy on OA\_UN, but much worse on the other criteria. This because Random strategy will take quite a few changed pixel pairs as training samples, which would mislead the training process of DSFA. In addition, the results of CVA are almost the same with results of Ground Truth strategy, which indicates that DSFA with a simple pre-detection step to generate training samples could also achieve the same valid performance with using the Ground Truth. And in the field of change detection, labeling ground truth are usually hard and time consuming in both research and practical problems. Therefore, CVA is taken as the pre-detection method in our proposed algorithm.
\begin{figure}[htbp]
	\centering
	{
		\includegraphics[scale=0.55]{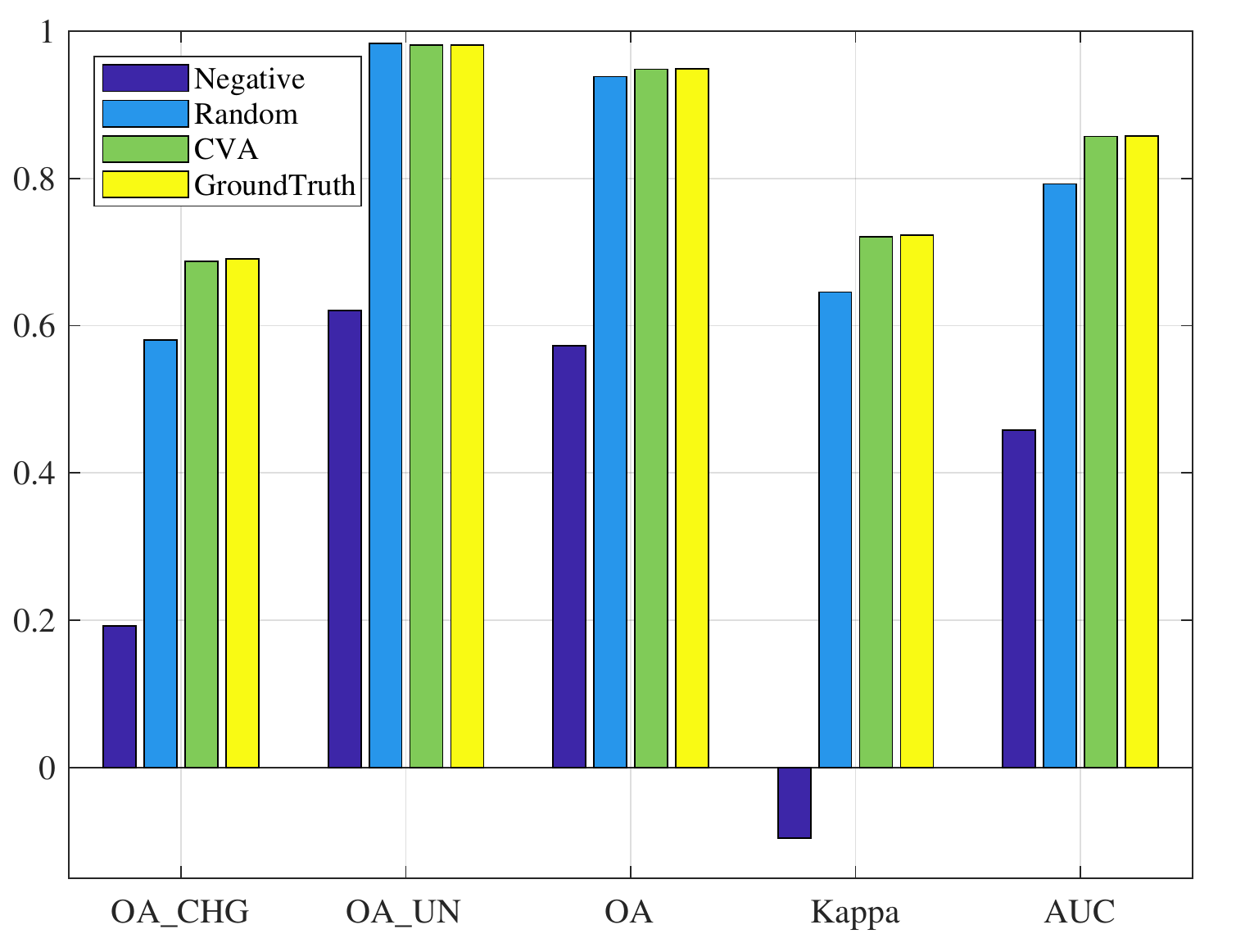}
		\caption{Comparison of different training sample selection strategies on River dataset.}
		\label{selection}}
\end{figure}
}

\section{Conclusion}\label{section5}

\par In this paper, we proposed a novel change detection algorithm called DSFA for multi-temporal remote sensing images. In the DSFA model, two deep networks are used to project the bi-temporal original input data into a new feature space. Then, SFA is used to extract the most invariant components of unchanged pixels and suppress them in changed regions to highlight changed components. We formulated the SFA process and loss function of DSFA model, and presented the derivation of computing gradient of loss. Our proposed algorithm is unsupervised, which means it doesn't need priori labeled pixels for the training process.
{\par We implemented our algorithm and performed experiments on two multi-spectral datasets and a public hyperspectral dataset. The visual and quantitative results have both shown that our method could outperform the other state-of-the-art methods, including other SFA-based and deep network algorithms.
\par Our proposed method currently focuses on differentiating the changed and unchanged regions in bi-temporal remote sensing imagery. The future work is required to explore DSFA's potential in detecting multi-classes changes. And in consideration of that SFA is originally designed for solving the problems of continuous signals, it will be promising to develop a specific DSFA model for change detection of sequent or video imagery.}

\appendices
\section{Derivation of Gradient of loss}\label{appendix}
\par Here we will present the detailed deduction process of computing the gradient of $\mathcal{L}(\theta_1,\theta_2)$ with respect to $\hat X_\phi$. Based on the reference \cite{IMM2012-03274}, we have the following equations.
\begin{equation}
	\frac{\partial tr(ABA^TC)}{\partial A}=CAB+C^TAB^T,
\end{equation}
\begin{equation}
	\frac{\partial(X^{-1})_{kl}}{\partial X_{ij}}=-(X^{-1})_{ki}(X^{-1})_{jl}.
\end{equation}
\par Based on (36) and the fact that $A_\phi$ and $B_\phi$ are both symmetric, we could obtain:
\begin{equation}
	\nabla_A=\frac{\partial \mathcal{L}(\theta_1,\theta_2)}{\partial A_\phi}=2B^{-1}_\phi A_\phi B^{-1}_\phi,
\end{equation}
\begin{equation}
	\begin{split}
		&\frac{\partial \mathcal{L}(\theta_1,\theta_2)}{\partial B^{-1}_\phi}=2A_\phi B^{-1}_\phi A_\phi \\ \Leftrightarrow (&\frac{\partial \mathcal{L}(\theta_1,\theta_2)}{\partial B^{-1}_\phi})_{kl}=2(A_\phi B^{-1}_\phi A_\phi)_{kl}.
	\end{split}
\end{equation}
\par Then, combining (37), $\nabla_B=\partial \mathcal{L}(\theta_1, \theta_2)/\partial B_\phi$ is calculated as the following equation:
\begin{equation}
	\begin{split}
		\nabla_B&=-2\sum_{kl}(A_\phi B^{-1}_\phi A_\phi)_{kl}(B^{-1}_\phi)_{ki}(B^{-1}_\phi)_{jl} \\ &=-2\sum_{kl}(B^{-1}_\phi)_{ik}(A_\phi B^{-1}_\phi A_\phi)_{kl}(B^{-1}_\phi)_{lj} \\ &=-2(B_\phi^{-1}A_\phi B_\phi^{-1}A_\phi B_\phi^{-1}).
	\end{split}
\end{equation}
\par We could expand the expression of $A_\phi$ out:
\begin{equation}
	\begin{split}
		A_\phi&=\Sigma_{XY}=\frac{1}{n}(\hat X_\phi-\hat Y_\phi)(\hat X_\phi-\hat Y_\phi)^T \\ &=\frac{1}{n}(\hat X_\phi \hat X_\phi^T+\hat Y_\phi \hat Y_\phi^T-\hat X_\phi \hat Y_\phi^T-\hat Y_\phi \hat X_\phi^T).
	\end{split}
\end{equation}
\par First, based on the derivation in the appendix of \cite{andrew2013deep}, we have:
\begin{equation}
	\begin{split}
		\frac{\partial (\hat X_\phi \hat X_\phi^T)^{ab}}{\partial \hat X_\phi^{ij}}&=\left\{
		\begin{array}{rcl}
			\frac{2}{n}(\hat X_\phi^{ij} -\frac{1}{n}\sum_k \hat X_\phi^{ik}), a=i, b=i      \\
			\frac{1}{n}(\hat X_\phi^{bj} -\frac{1}{n}\sum_k \hat X_\phi^{bk}), a=i, b\neq i  \\
			\frac{1}{n}(\hat X_\phi^{aj} -\frac{1}{n}\sum_k \hat X_\phi^{ak}), a \neq i, b=i \\
			0, a\neq i, b \neq i                                                             \\
		\end{array} \right.\\
		&=\frac{1}{n}(\xi_{(a=i)}\hat X_\phi^{bj}+\xi_{(b=i)}\hat X_\phi^{aj}).
	\end{split}
\end{equation}
\par Also,
\begin{equation}
	\frac{\partial(\hat X_\phi \hat Y_\phi^T)^{ab}}{\partial \hat X_\phi^{ij}}=\frac{1}{n}(\hat Y_\phi^{bj}-\frac{1}{n}\sum_k\hat Y_\phi^{bk})=\frac{1}{n}\xi_{(a=i)}\hat Y_\phi^{bj}.
\end{equation}
\par Integrating (42) and (43) into (41):
\begin{equation}
	\begin{split}
		\frac{\partial A_\phi^{ab}}{\partial \hat X_\phi^{ij}}&=\frac{\partial(\hat X_\phi \hat X_\phi^T)^{ab}}{\partial \hat X_\phi^{ij}}-\frac{\partial(\hat Y_\phi \hat X_\phi^T)^{ab}}{\partial \hat X_\phi^{ij}}-\frac{\partial(\hat X_\phi \hat Y_\phi^T)^{ab}}{\partial \hat X_\phi^{ij}} \\
		&=\frac{1}{n}(\xi_{(a=i)}\hat X_\phi^{bj}+\xi_{(b=i)}\hat X_\phi^{aj})\\
		&-\frac{1}{n}(\xi_{(b=i)}\hat Y_\phi^{aj}+\xi_{(a=i)}\hat Y_\phi^{bj}).
	\end{split}
\end{equation}
\par Similarly, with respect to $B_\phi$, we have:
\begin{equation}
	\begin{split}
		\frac{\partial B_\phi^{ab}}{\partial \hat X_\phi^{ij}}&=\frac{\partial \Sigma_{XX}^{ab}+\partial \Sigma_{YY}^{ab}}{2\partial \hat X_\phi^{ij}}\\
		&=\frac{1}{2n}(\xi_{(a=i)}\hat X_\phi^{bj}+\xi_{(b=i)}\hat X_\phi^{aj}).
	\end{split}
\end{equation}
\par Putting (44) and (45) together, the gradient of $\mathcal{L}(\theta_1, \theta_2)$ with respect to $\hat X_\phi^{ij}$ is then computed as:
\begin{equation}
	\begin{split}
		\frac{\partial \mathcal{L}(\theta_1, \theta_2)}{\partial \hat X_\phi^{ij}}&=\sum_{ab}\nabla_A^{ab}\frac{\partial A_\phi^{ab}}{\partial \hat X_\phi^{ij}}+\sum_{ab}\nabla_B^{ab}\frac{\partial B_\phi^{ab}}{\partial \hat X_\phi^{ij}}\\
		&=\frac{1}{n}(\sum_b \nabla_A^{ib}\hat X_\phi^{bj}+\sum_a \nabla_A^{ai}\hat X_\phi^{aj})\\
		&-\frac{1}{n}(\sum_b \nabla_A^{ib}\hat Y_\phi^{bj}+\sum_a \nabla_A^{ai}\hat Y_\phi^{aj})\\
		&+\frac{1}{2n}(\sum_b \nabla_B^{ib}\hat X_\phi^{bj}+\sum_a \nabla_B^{ai}\hat X_\phi^{aj})\\
		&=\frac{1}{n}(\nabla_A\hat X_\phi+\nabla_A^T\hat X_\phi-\nabla_A\hat Y_\phi-\nabla_A^T\hat Y_\phi)_{ij}\\
		&+\frac{1}{2n}(\nabla_B\hat X_\phi+\nabla_B^T\hat X_\phi)_{ij}.
	\end{split}
\end{equation}
\par Obviously, $\nabla_A$ and $\nabla_B$ are both symmetric matrices. Therefore,
\begin{equation}
	\frac{\partial\mathcal{L}(\theta_1,\theta_2)}{\partial\hat X_\phi^{ij}}=\frac{2}{n}(\nabla_A\hat X_\phi-\nabla_A\hat Y_\phi)_{ij}+\frac{1}{n}(\nabla_B\hat X_\phi)_{ij}.
\end{equation}
\par Finally, we could obtain the gradient of $\mathcal{L}(\theta_1, \theta_2)$ with respect to $\hat X_\phi$:
\begin{equation}
	\frac{\partial\mathcal{L}(\theta_1,\theta_2)}{\partial\hat X_\phi}=\frac{2}{n}(\nabla_A\hat X_\phi-\nabla_A\hat Y_\phi)+\frac{1}{n}\nabla_B\hat X_\phi.
\end{equation}



\ifCLASSOPTIONcaptionsoff
	\newpage
\fi

\bibliographystyle{IEEEtran}
\bibliography{IEEEabrv,final.bbl}
%




\end{document}